\documentclass[10pt, letterpaper, journal, twoside]{IEEEtran}

\IEEEoverridecommandlockouts    
   
\newcommand\highlightReference[1]{%
  \expandafter\newcommand\csname highlightReference-#1\endcsname{}%
}
\let\oldbibitem\bibitem
\def\bibitem#1#2\par{%
  \expandafter\ifx\csname highlightReference-#1\endcsname\relax
    \oldbibitem{#1}#2\par
  \else
    \oldbibitem{#1}\highlight{#2}\par
  \fi
}
\usepackage{color}
\newcommand\highlight[1]{\textcolor{red}{#1}}

\usepackage{graphics} 
\usepackage{cite}
\usepackage[T1]{fontenc}
\usepackage{amsmath, bm} 
\usepackage{amssymb}  

\usepackage{hyperref}
\hypersetup{
    colorlinks=true,
    linkcolor=blue,
    filecolor=magenta,      
    urlcolor=blue,
}

\usepackage{float}
\usepackage{cancel}

\usepackage{enumitem}

\usepackage{numprint}
\usepackage{comment}

\usepackage{color, colortbl}
\definecolor{LightCyan}{rgb}{0.88,1,1}
\usepackage{multirow}
\usepackage{siunitx}
\DeclareSIUnit\cell{cell}
\DeclareSIUnit\cells{cells}
\DeclareSIUnit\trees{trees}

\newif\ifincludeFigures
\includeFigurestrue 

\usepackage{float}
\makeatletter
\def\fps@figure{t}
\def\fps@table{t}
\makeatother

\usepackage{enumitem} 

\usepackage{wrapfig}
\usepackage[table]{xcolor}
\usepackage{amssymb}
\usepackage{tikz}
\usepackage{pgfplots}
\usetikzlibrary{shapes,arrows}
\usepackage{verbatim}
\usetikzlibrary{positioning}
\usepackage[normalem]{ulem} 
\renewcommand{\sout}[1]{} 
\usepackage{tabstackengine}
\usepackage{algorithm}
\usepackage{algorithmicx}
\usepackage[end]{algpseudocode} 
\usepackage{etoolbox}

\usepackage[font=small]{caption}
\usepackage{subcaption}

\usepackage{pgfplots}
\usepackage{grffile}
\usetikzlibrary{plotmarks}
\usetikzlibrary{arrows.meta}
\usepgfplotslibrary{patchplots}
\usepackage{amsmath}
\usepackage{scalerel}
\newcommand\Tau{\footnotesize \scalerel*{\tau}{T}}

\usepackage{bbding} 

\usepackage{colortbl}
\usepackage[table]{xcolor}

\usepackage{hhline}
\usepackage{makecell}

\newcommand{\ihab}[1]{{\textcolor{black}{#1}}}
\newcommand{\junhong}[1]{{\textcolor{black}{#1}}}

\usepackage{eso-pic}
\newcommand\AtPageUpperMyright[1]{\AtPageUpperLeft{%
 \put(\LenToUnit{0.5\paperwidth},\LenToUnit{-1.5cm}){%
     \parbox{0.5\textwidth}{\raggedright\fontsize{11}{11}\selectfont #1}}%
 }}%
\newcommand{\conf}[1]{%
\AddToShipoutPictureBG*{%
\AtPageUpperMyright{#1}
}
}

\title{\LARGE \bf Towards Efficient MPPI Trajectory Generation with Unscented Guidance: U-MPPI Control Strategy}

\conf{\hspace*{-9cm}\textcolor{gray}{This paper has been accepted for publication in the \textbf{IEEE Transactions on Robotics} \textcolor{blue}{\href{https://ieeexplore.ieee.org/xpl/RecentIssue.jsp?punumber=8860}{(T-RO)}}, December 2024.}}

\author{Ihab S. Mohamed$^{1}$, Junhong Xu$^{1}$, Gaurav S Sukhatme$^{2}$, and Lantao Liu$^{1}$
\thanks{$^{1}$Ihab S. Mohamed, Junhong Xu, and Lantao Liu are with the Luddy School of Informatics, Computing, and Engineering, Indiana University, Bloomington, IN 47408 USA (e-mail: {\tt\small \{mohamedi, xu14, lantao\}@iu.edu})}
\thanks{$^{2}$Gaurav S. Sukhatme is with the Department of Computer Science, University of Southern California, Los Angeles, CA 90089, USA (e-mail: {\tt\small 
gaurav@usc.edu)}}
\thanks{This work is supported by the National Science Foundation (NSF) under grant numbers 2006886 and 2047169.}
\thanks{Corresponding author: Lantao Liu}
}%
\definecolor{applegreen}{rgb}{0.8, 1, 0.0}
\definecolor{LightCyan}{rgb}{0.88,1,1}
\definecolor{atomictangerine}{rgb}{1.0, 0.6, 0.4}
\definecolor{amber}{rgb}{1.0, 0.75, 0.0}
\definecolor{aqua}{rgb}{0.0, 1.0, 1.0}
\definecolor{almond}{rgb}{0.94, 0.87, 0.8}
\definecolor{aquamarine}{rgb}{0.5, 1.0, 0.83}
\definecolor{babyblue}{rgb}{0.54, 0.81, 0.94}
\definecolor{babyblueeyes}{rgb}{0.63, 0.79, 0.95}
\definecolor{asparagus}{rgb}{0.53, 0.66, 0.42}
\definecolor{auburn}{rgb}{0.43, 0.21, 0.1}
\definecolor{brilliantlavender}{rgb}{0.96, 0.73, 1.0}
\definecolor{bittersweet}{rgb}{1.0, 0.44, 0.37}
\definecolor{blue-violet}{rgb}{0.54, 0.17, 0.89}
\definecolor{capri}{rgb}{0.0, 0.75, 1.0}
\definecolor{celadon}{rgb}{0.67, 0.88, 0.69}
\definecolor{darkcyan}{rgb}{0.0, 0.55, 0.55}
\definecolor{deepskyblue}{rgb}{0.0, 0.75, 1.0}
\definecolor{dogwoodrose}{rgb}{0.84, 0.09, 0.41}

\begin{document}


\maketitle

\global\csname @topnum\endcsname 0
\global\csname @botnum\endcsname 0


\thispagestyle{empty}
\pagestyle{empty}

\begin{abstract}
The classical Model Predictive Path Integral (MPPI) control framework\ihab{, while effective in many applications,} lacks reliable safety features \ihab{\sout{since it relies} due to its reliance} on a \textit{risk-neutral} trajectory evaluation technique, which can present challenges for safety-critical applications such as autonomous driving. Furthermore, \ihab{\sout{if} when} the majority of MPPI sampled trajectories concentrate in high-cost regions, it may generate an \textit{infeasible} control sequence.
To address this challenge, we propose the U-MPPI control strategy, a novel methodology that can effectively manage system uncertainties while integrating a more efficient trajectory sampling strategy. The core concept is to leverage the Unscented Transform (UT) to propagate not only the mean but also the covariance of the system dynamics, going beyond the traditional MPPI method. As a result, it introduces a novel and more efficient trajectory sampling strategy, significantly enhancing state-space exploration and ultimately reducing the risk of being trapped in local minima.
Furthermore, by leveraging the uncertainty information provided by UT, we incorporate a \textit{risk-sensitive} cost function that explicitly accounts for risk or uncertainty throughout the trajectory evaluation process, resulting in a more resilient control system capable of handling uncertain conditions.
By conducting extensive simulations of 2D aggressive autonomous navigation in both known and unknown cluttered environments, we verify the efficiency and robustness of our proposed U-MPPI control strategy compared to the baseline MPPI. 
We further validate the practicality of U-MPPI through real-world demonstrations in unknown cluttered environments, showcasing its superior ability to incorporate both the UT and local costmap into the optimization problem without introducing additional complexity.
\end{abstract}
\begin{IEEEkeywords}
Autonomous vehicle navigation, MPPI, unscented transform, occupancy grid map path planning.
\end{IEEEkeywords}
\section*{Supplementary Material}
A video showcasing both the simulation and real-world results, including the behavior of the vanilla MPPI, is provided in the supplementary video: \url{https://youtu.be/1xsh4BxIrng}\\
Moreover, the GPU implementation of our proposed U-MPPI algorithm is available at: \url{https://github.com/IhabMohamed/U-MPPI} 

\section{Introduction and Related Work}\label{Introduction}
\begin{figure}%
    \centering
        \includegraphics[scale=0.68]{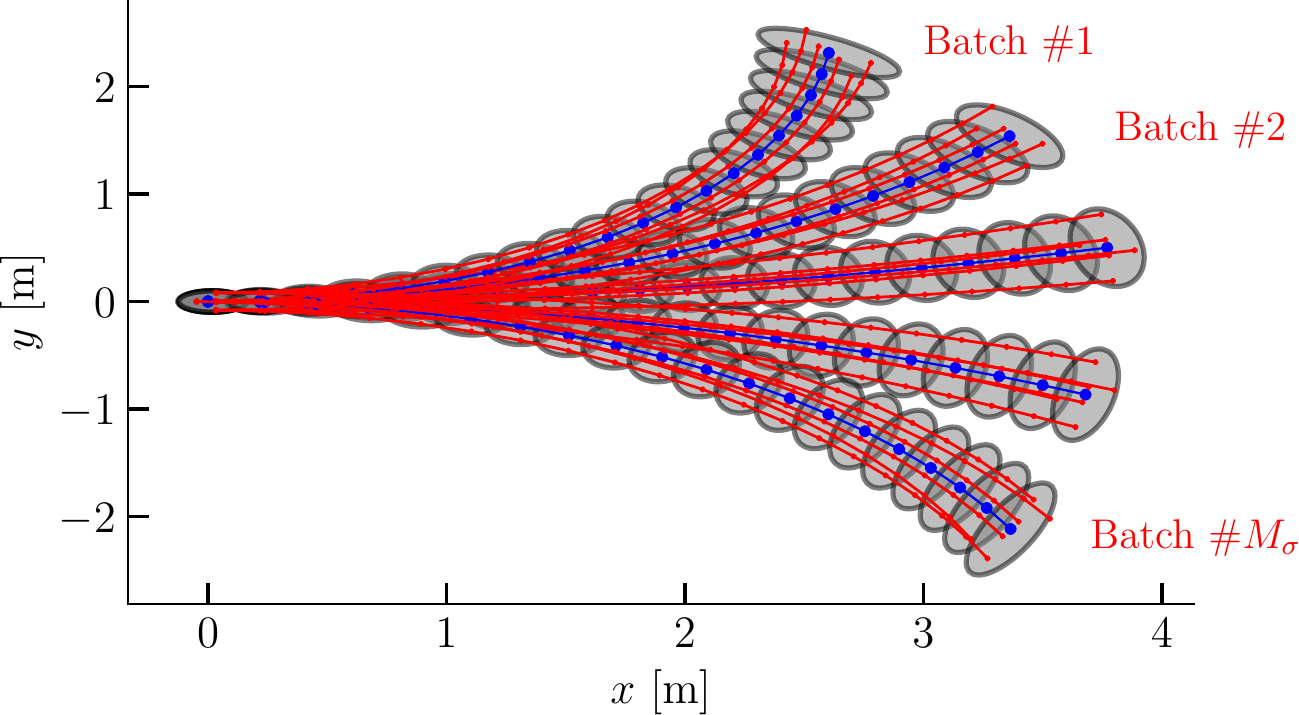}
    \caption{Our proposed sampling strategy, for a ground vehicle model, under the U-MPPI control strategy based on unscented transform; such a sampling strategy propagates both the mean $\bar{\mathbf{x}}_k$ (blue dots) and covariance $\mathbf{\Sigma}_k$ (gray ellipses) of the state vector at each time-step $k$; to generate $M$ sampled trajectories, we propagate $M_\sigma$ sets of batches, where each batch contains $n_\sigma$ trajectories corresponding to the $n_\sigma$ sigma points, where $M=n_\sigma M_\sigma$, $n_\sigma = 2n_x +1$, and red lines refer to $2n_x$ sigma-point trajectories surrounding the nominal trajectories (blue lines); for our validation-used robot, $n_x=3$.
    }%
    \label{fig:vis-ut-sampling-strategy}%
\end{figure}
 \IEEEPARstart{P}{lanning} and controlling autonomous vehicles under uncertainty is a partially-solved yet highly challenging problem in robotics. The uncertainty or risk can arise from multiple sources, such as the vehicle's dynamics, the dynamics of other objects in the environment, the presence of unexpected obstacles, and the accuracy of the sensors used to perceive the environment. These factors can lead to unpredictable vehicle behavior, making it challenging to anticipate its movements and actions.
 Therefore, it is crucial for motion planning and control algorithms to account for the uncertainties of the system's states, sensing, and control actions. This eventually enables autonomous vehicles to adapt to unexpected environmental changes and make reliable decisions in real-time \cite{kurniawati2011motion}.

Model Predictive Control (MPC), also referred to as \textit{receding-horizon} optimal control has been introduced as an effective solution for improving system safety and managing random disturbances and uncertainties, owing to its flexibility  and ability to handle system constraints and nonlinearities while optimizing system performance. 
It plans a sequence of optimal control inputs over a finite time-horizon by repeatedly solving the optimal control problem using \junhong{the \textit{receding-horizon} principle (i.e., optimal control solvers~\cite{mattingley2011receding})}, with the first control input applied to the system.
There are two main categories of MPC frameworks that can handle various forms of uncertainty: Robust MPC (RMPC) and Stochastic MPC (SMPC). RMPC considers worst-case scenarios of uncertainty to ensure stability and constraint satisfaction, while SMPC takes into account expected costs and probabilistic constraints (e.g., chance constraints) to optimize performance under uncertain conditions.
\ihab{\sout{RMPC is known to generate the safest solutions among MPC categories due to its incorporation of worst-case scenarios into the optimization problem and its emphasis on minimizing worst-case objective functions.
However, it may lead to excessively conservative control actions, resulting in low system performance}
In RMPC, a \textit{minimax} optimization problem is solved to ensure the optimal performance of the control strategy under worst-case scenarios while guaranteeing the satisfaction of constraints for all possible uncertainties \cite{lofberg2003minimax}. This method is known for generating the safest solutions among MPC categories due to its incorporation of worst-case scenarios into the optimization process. Nevertheless, its emphasis on minimizing worst-case objective functions often results in complex, intractable optimization problems, leading to overly conservative control actions and reduced system efficiency~\cite{bemporad2007robust, mesbah2016stochastic}.} 

On the other hand, despite the superior ability of SMPC to leverage the probabilistic nature of uncertainties, many SMPC approaches have performance limitations, including being tailored to specific forms of stochastic noise, requiring dynamics linearization to ensure a \textit{real-time} performance and implementation of the control strategy, and employing chance constraints that may not accurately reflect the severity of constraint violations or potential accidents and can be computationally demanding to evaluate, particularly for complex or high-dimensional probability distributions \cite{du2011probabilistic, knudsen2018stochastic, hewing2018stochastic}.
Additionally, SMPC has a further limitation in that it relies on a \textit{risk-neutral} expectation to predict future uncertain outcomes, which may not be reliable in the case of a tail-end probability event (i.e., a rare event with a low probability of occurring) actually happening \cite{ sajadi2017risk}.
To address the challenges posed by uncertainties, Risk-Sensitive MPC (RSMPC) approaches have gained traction in recent years, thanks to their ability to balance the benefits and drawbacks of robust and stochastic MPC methods. By integrating the concept of \textit{risk measures} or \textit{risk metrics} into the optimization problem, RSMPC can evaluate the impact of uncertainty and adjust responses accordingly to different levels of uncertainty \cite{yang2015risk, hyeon2020fast, schuurmans2020learning}.

The Model Predictive Path Integral (MPPI) framework, a type of SMPC method, has emerged as a promising control strategy for complex robotics systems with stochastic dynamics and uncertainties \cite{williams2017model, williams2018information, mohamed2020model, pravitra2020L, mohamed2021sampling}.
Such a method solves the stochastic optimal control problem in a \textit{receding-horizon} control setting by: (i) leveraging Monte Carlo simulation to rollout real-time simulated trajectories propagated from the system dynamics, (ii) evaluating these trajectories, (iii) computing the optimal control sequence by taking the weighted average of the costs of the sampled trajectories, and (iv) applying the first control input to the system while using the remaining control sequence to warm-start the optimization in the next time-step, enabling the method to solve the optimization problem effectively \cite{williams2017model, kazim2024recent}. 
MPPI stands out among alternative MPC methods due to its attractive features, such as being a sampling-based and derivative-free optimization method, not relying on assumptions or approximations of objective functions and system dynamics, being effective for highly dynamic systems, and benefiting from parallel sampling and the computational capabilities of Graphics Processing Units (GPUs) to achieve optimized and \textit{real-time} performance \cite{mohamed2021mppi}.\footnote{It is worth noting that a CPU-based MPPI, optimized using vectorization and tensor operations, is now available. See the link for further information: \url{https://github.com/artofnothingness/mppic}} \junhong{Due to these attractive features, even the simplified Predictive Sampling method introduced in \cite{howell2022predictive}, which can variously be described as MPPI with infinite temperature, can achieve competitive performance with more established gradient-based algorithms.}


While MPPI has appealing characteristics, it may also pose challenges in practice. One particular concern is that, much like any sampling-based optimization algorithm, it could generate an \textit{infeasible} control sequence if all the resulting MPPI sampled trajectories are concentrated in a high-cost region, which may lead to violations of system constraints or a higher likelihood of being trapped in local minima \cite{yin2022trajectory, mohamed2022autonomous}.
In \cite{williams2018robust}, Tube-MPPI was proposed as a solution to alleviate the situation by incorporating an iterative Linear Quadratic Gaussian (iLQG) controller (which, unfortunately, requires the linearization of dynamics) as an ancillary controller to track the MPPI-generated nominal trajectory. 
Similarly, in \cite{pravitra2020L}, an augmented version of MPPI is employed, which includes a nonlinear $\mathcal{L}_1$ adaptive controller to address model uncertainty. 
\ihab{In \cite{mohamed2023gp}, the authors propose the GP-MPPI control strategy to ensure efficient and safe navigation in unknown and complex environments while avoiding local minima. This strategy combines MPPI with a Sparse Gaussian Process (SGP)-based local perception model, incorporating online learning to effectively explore the surrounding navigable space. Meanwhile, in \cite{rastgar2024priest}, the authors introduce Projection Guided Sampling Based Optimization (PRIEST), a new optimizer designed to push the infeasible sampled trajectories of MPPI towards feasible regions, thereby reducing the risk of local minima and infeasible control sequences.}

\ihab{Recently, various sampling techniques have been introduced to enhance the performance of MPPI \cite{yin2022trajectory, mohamed2022autonomous, gandhi2023safe, asmar2023model, qu2023rl, yan2024output, trevisan2024biased, yin2023shield}.}
\ihab{To name a few examples,} \cite{yin2022trajectory} enhances the MPPI algorithm by incorporating the covariance steering (CS) principle, while \cite{mohamed2022autonomous} proposes sampling trajectories from a product of normal and log-normal distributions (NLN mixture), instead of solely using a Gaussian distribution. 
\ihab{In \cite{gandhi2023safe}, a novel method embeds safety filters during forward sampling to ensure the safe operation of robotic systems within multiple constraints. 
Additionally, \cite{asmar2023model} improves MPPI performance by integrating adaptive importance sampling into the original importance sampling method, while \cite{trevisan2024biased} proposes a new importance sampling scheme that combines classical and learning-based ancillary controllers to improve efficiency.}
These methods result in more efficient trajectories than the vanilla MPPI, leading to better exploration of the system's state-space and reducing the risk of encountering local minima.
\ihab{\cite{yin2023shield} presents an algorithm, called Shield-MPPI, that enhances the robustness and computational efficiency of MPPI using Control Barrier Functions (CBFs) to ensure safety and reduce the need for extensive computational resources, making it suitable for real-time planning and deployment on hardware without GPUs.}

Another constraint of MPPI is its inability to explicitly incorporate risk levels during planning due to its \textit{risk-neutral} technique in evaluating sampled trajectories during the optimization process, making it challenging to achieve the desired balance between risk and robustness.
Additionally, the MPPI optimization problem concentrates solely on minimizing the objective function, which is influenced by a minor perturbation injected into the control input, without explicitly considering any uncertainties or risks associated with the system dynamics or the environment.
This can eventually lead to sub-optimal or overly aggressive control actions, as MPPI may select trajectories that appear to have a low expected cost but may actually be riskier or less robust in practice.
Consequently, MPPI cannot guarantee safety when environmental conditions change, which limits its applicability for safety-critical applications such as autonomous driving. \cite{wang2021adaptive} \ihab{introduces a general framework for optimizing Conditional Value-at-Risk (CVaR) under uncertain dynamics, parameters, and initial conditions by incorporating additional samples from sources of uncertainty.
}
The Risk-aware MPPI (RA-MPPI) algorithm \cite{yin2023risk} is a more recent approach that addresses this issue by utilizing CVaR to generate risk-averse controls that evaluate real-time risks and account for systematic uncertainties. 
However, such a method employs Monte-Carlo sampling to estimate the CVaR, which can be computationally intensive and time-consuming since generating a large number of random samples is required for accurate estimation using Monte Carlo methods. 

\ihab{In contrast to existing solutions aimed at mitigating the shortcomings of MPPI, the aim of our proposed solution is to effectively consider the uncertainties of system states, sensing, and control actions, in addition to integrating a novel and more efficient trajectory sampling strategy.
To this end,} we introduce the U-MPPI control strategy, a novel methodology that enhances the classical MPPI algorithm by combining the Unscented Transform (UT) with standard optimal control theory
(also known as unscented guidance \cite{ross2015unscented}) to effectively manage system uncertainties \ihab{and improve trajectory sampling efficiency.} Such a control strategy leverages the UT for two purposes: regulating the propagation of the dynamical system and proposing a new state-dependent cost function formulation that incorporates uncertainty information. 
To the best of the authors' knowledge, the proposed control strategy has not been previously explored in the literature\ihab{, presenting a unique integration of the UT with the MPPI control framework. }   
 In summary, the contributions of this work can be summarized as follows:
\begin{enumerate}
     \item 
     \ihab{\sout{While vanilla MPPI variants propagate only the mean value of the system dynamics, as depicted in Fig.~\ref{fig:mppi_dynamical_propagation}, we propose a novel trajectory sampling technique equipped with a more effective sampling distribution policy based on UT; this technique utilizes UT to propagate both the mean and covariance of the system dynamics at each time-step, as demonstrated in Figs.~\ref{fig:vis-ut-sampling-strategy} and \ref{fig:umppi_dynamical_propagation}, and explained in detail in Section~\ref{Unscented-Based Sampling Strategy}; by doing so, our new sampling method achieves significantly better exploration of the state-space of the given system and reduces the risk of getting trapped in local minima.} While vanilla MPPI variants propagate only the mean value of the system dynamics, we propose a novel trajectory sampling technique utilizing the UT to propagate both the mean and covariance of the system dynamics, significantly enhancing state-space exploration and reducing the risk of local minima.}
    \item Then, by utilizing the propagated uncertainty information (i.e., state covariance matrix), we introduce a \textit{risk-sensitive} cost function that explicitly considers risk or uncertainty during the trajectory evaluation process, leading to a safer and more robust control system, especially for safety-critical applications.\ihab{\sout{, as discussed in Section~\ref{Risk-Sensitive Cost}.}}
    \item \ihab{We provide a detailed analysis of the risk-sensitive function and its derivation, offering an in-depth understanding of the U-MPPI control strategy's behavior under different risk conditions, which elucidates how it adapts and performs in various scenarios, thereby enhancing its robustness and safety in uncertain environments.}
    \ihab{\sout{In Sections \ref{Simulation Details and Results} and \ref{Real-World Demonstration}, we validate the effectiveness of our U-MPPI control strategy for aggressive collision-free navigation in both known and unknown cluttered environments using both intensive simulations and real-world experiments; by comparing it with the baseline MPPI, we demonstrate its superiority in producing more efficient trajectories that explore the state-space of the system better, resulting in higher success and task completion rates, reducing the risk of getting trapped in local minima, and ultimately leading the robot to find feasible trajectories that avoid collisions.}}
\end{enumerate}
\ihab{
Through extensive simulations and real-world experiments, we demonstrate the effectiveness of our U-MPPI control strategy for aggressive collision-free navigation in both known and unknown cluttered environments, highlighting its superiority over the baseline MPPI control framework by producing more efficient trajectories, achieving higher success and task completion rates, reducing the risk of local minima, and ultimately enabling the robot to find feasible collision-free trajectories.}

\ihab{ The rest of the paper is organized as follows. Section~\ref{Stochastic Optimal Control} addresses the problem statement of stochastic optimal control and provides an overview of the vanilla MPPI control strategy. Section~\ref{Unscented Optimal Control} introduces the unscented transform and its application in generating robust control strategies by combining it with standard optimal control theory to manage system uncertainties. Section~\ref{U-MPPI Control Strategy} details the proposed U-MPPI control strategy, including the unscented-based sampling strategy, the \textit{risk-sensitive} cost function, and its impact on the controller's behavior, and explains the real-time control algorithm.
Following this, Section~\ref{Simulation Details and Results} presents simulation results comparing the performance of the U-MPPI control strategy with the baseline MPPI in various scenarios, highlighting improvements in trajectory generation and robustness. Subsequently, Section~\ref{Real-World Demonstration} demonstrates the practicality of the U-MPPI control strategy through real-world experiments in unknown cluttered environments, showcasing its effectiveness and adaptability.
Finally, Section~\ref{sec:conclusion} concludes the study and proposes potential future research directions.}

\section{Stochastic Optimal Control}\label{Stochastic Optimal Control}
This section aims to establish the problem statement of stochastic optimal control and present a concise overview of MPPI as a potential solution to address this problem. 
\subsection{Problem Formulation}\label{Problem Formulation}
Within the context of discrete-time stochastic systems, let us consider the system state $\mathbf{x}_k \in \mathbb{R}^{n_x}$, the control input $\mathbf{u}_k \in \mathbb{R}^{n_u}$, and the underlying non-linear dynamics 
\begin{equation}\label{eq: underlying non-linear dynamics}
   \mathbf{x}_{k+1}=f\left(\mathbf{x}_{k},\mathbf{w}_{k}\right).    
\end{equation}
The actual (i.e., disturbed) control input, $\mathbf{w}_{k}$, is represented as $\mathbf{w}_{k} = \mathbf{u}_{k}+\delta \mathbf{u}_{k}\sim \mathcal{N}(\mathbf{u}_k, \Sigma_{\mathbf{u}})$, where $\delta \mathbf{u}_{k} \sim \mathcal{N}(\mathbf{0}, \Sigma_{\mathbf{u}})$ is a zero-mean Gaussian noise with covariance $\Sigma_{\mathbf{u}}$ which represents the injected disturbance into the control input.
Within a finite time-horizon $N$, we denote the control sequence
$\mathbf{U} = \left[\mathbf{u}_{0}, \mathbf{u}_{1}, \dots,\mathbf{u}_{N-1}\right]^{\top} \in \mathbb{R}^{n_u N}$ and the corresponding state trajectory $\mathbf{x} = \left[\mathbf{x}_{0}, \mathbf{x}_{1}, \dots, \mathbf{x}_{N}\right]^{\top} \in \mathbb{R}^{n_x (N+1)}$. Moreover, 
let $\mathcal{X}^d$ denote the $d$ dimensional space with  $\mathcal{X}_{rob}\left(\mathbf{x}_{k}\right) \subset \mathcal{X}^d$ and $\mathcal{X}_{o b s} \subset \mathcal{X}^d$ represent the area occupied by the robot and obstacles, respectively.
In this scenario, the objective of the stochastic optimal control problem is to find the optimal control sequence, $\mathbf{U}$, that generates a collision-free trajectory, guiding the robot from its initial state, $\mathbf{x}_s$, to the desired state, $\mathbf{x}_f$, under the minimization of the cost function, $J$, subject to specified constraints.
The optimization problem at hand can be formulated using the vanilla MPPI control strategy as
\begin{subequations}
\begin{align}
\min _{\mathbf{U}} \,\; J (\mathbf{x},\mathbf{u}) &=  \mathbb{E}\left[\phi\left(\mathbf{x}_{N}\right)+\!\!\sum_{k=0}^{N-1}\!\!\left(\!q\!\left(\mathbf{x}_{k}\right)+\frac{1}{2} \mathbf{u}_{k}^{\top}  R \mathbf{u}_{k}\!\!\right)\!\!\right]\!, \label{eq:2a}\\
\text {s.t.} 
\quad & \mathbf{x}_{k+1}=f\left(\mathbf{x}_{k}, \mathbf{w}_{k}\right), \delta \mathbf{u}_{k} \sim \mathcal{N}(\mathbf{0}, \Sigma_{\mathbf{u}}), \label{eq:2b}\\
& \mathcal{X}_{rob}\left(\mathbf{x}_{k}\right) \cap \mathcal{X}_{obs}=\emptyset, \;h(\mathbf{x}_k, \mathbf{u}_k) \leq 0, \label{eq:2c}\\
& \mathbf{x}_0 = \mathbf{x}_s, \;\mathbf{u}_{k} \in \mathbb{U},\; \mathbf{x}_{k} \in \mathbb{X}, 
\end{align}
\label{eq:2}
\end{subequations}
where $R\in \mathbb{R}^{n_u \times n_u}$ is a positive-definite control weighting matrix, $\mathbb{U}$ denotes the set of admissible control inputs, and $\mathbb{X}$ denotes the set of all possible states $\mathbf{x}_k$; the state terminal cost function, $\phi\left(\mathbf{x}_N\right)$, and the running cost function, $q\left(\mathbf{x}_{k}\right)$, can be defined as arbitrary functions,
offering a more flexible and dynamic approach to cost modeling that can be adapted to meet the specific requirements of the system being controlled.
\ihab{The function \(h(\mathbf{x}_k, \mathbf{u}_k)\) represents the state and control constraints that need to be satisfied at each time step. For instance, \(h(\mathbf{x}_k, \mathbf{u}_k)\) can specify speed limits, actuator constraints, or state constraints such as position or orientation bounds. These constraints are met when \( h(\mathbf{x}_k, \mathbf{u}_k) \leq 0 \), ensuring that the system operates within safe and feasible regions. Additionally, to handle collision avoidance, the constraint \(\mathcal{X}_{\text{rob}}(\mathbf{x}_k) \cap \mathcal{X}_{\text{obs}} = \emptyset\) ensures that the robot does not collide with any obstacles.
}

\subsection{Overview of MPPI Control Strategy}\label{Overview of MPPI Control Strategy}
MPPI solves the optimization problem defined in (\ref{eq:2}) by minimizing the objective function $J$ (\ref{eq:2a}), taking into account the system dynamics (\ref{eq:2b}) and constraints, including collision avoidance and control constraints, detailed in (\ref{eq:2c}).
To this end, at each \ihab{\sout{time-step} control loop interval} $\Delta t$, MPPI employs the Monte Carlo simulation to sample thousands of \textit{real-time} simulated trajectories, represented by $M$, propagated from the underlying system dynamics, as illustrated in Fig.~\ref{fig:mppi_dynamical_propagation}. 
Subsequently, within the time-horizon $N$, the \textit{cost-to-go} of each trajectory $\tau_m$ can be evaluated as
\begin{equation}\label{cost-to-go-mppi}
 \tilde{S}\left(\tau_m \right) =\phi\left(\mathbf{x}_N\! \right) + \! \! \sum_{k=0}^{N-1} \!\tilde{q}\left(\mathbf{x}_{k}, \mathbf{u}_{k}, \delta \mathbf{u}_{k,m}\right), \forall m \in \! \{1, \cdots\!, M\},
\end{equation}
where $\phi(\mathbf{x}_N)$ refer to the terminal state cost, while the instantaneous running cost $\tilde{q}\left(\mathbf{x}_k, \mathbf{u}_k, \delta \mathbf{u}_k\right)$ encompasses both the state-dependent running cost $q\left(\mathbf{x}_{k}\right)$ and the quadratic control cost \ihab{\sout{$q\left(\mathbf{u}_{k}, \delta \mathbf{u}_{k} \right)$} $q_{\mathbf{u}}\left(\mathbf{u}_{k}, \delta \mathbf{u}_{k} \right)$} and is formulated as 
\begin{equation}\label{eq:cost-to-go}
\tilde{q} 
\!= \!
\underbrace{
\vphantom{
\gamma_{\mathbf{u}} \delta \mathbf{u}_{k}^{\top}  R \delta \mathbf{u}_{k}+\mathbf{u}_{k}^{\top}  R \delta \mathbf{u}_{k}+\frac{1}{2} \mathbf{u}_{k}^{\top}  R \mathbf{u}_{k}}
q\left(\mathbf{x}_{k}\!\right)}_{\text{\color{black}{\textit{State-dep.}}}} 
\!+ 
\underbrace{\gamma_{\mathbf{u}} \delta \mathbf{u}_{k,m}^{\top}  R \delta \mathbf{u}_{k,m}\!+ \mathbf{u}_{k}^{\top}  R \delta \mathbf{u}_{k,m}\!+ \frac{1}{2} \mathbf{u}_{k}^{\top}  R \mathbf{u}_{k}}_{\text{\color{black}{\textit{\ihab{$q_\mathbf{u}\left(\mathbf{u}_{k}, \delta \mathbf{u}_{k} \right)$}: Quadratic Control Cost}}}},
\end{equation}
where $\gamma_\mathbf{u} = \frac{\nu -1}{2\nu} \ihab{ \; \in \mathbb{R}_{\geq 0}}$, and \ihab{\sout{$\nu \in \mathbb{R}^{+}$}} $\ihab{\nu \geq 1 }$ determines the level of aggressiveness in exploring the state-space.

As stated in \cite{williams2017model}, the vanilla MPPI algorithm updates the optimal control sequence $\left\{\mathbf{u}_{k}\right\}_{k=0}^{N-1}$ by considering a weighted average cost from all of the simulated trajectories; mathematically,
this control sequence can be expressed as
\begin{equation}\label{eq:mppi_optimal-control}
 \mathbf{u}_{k} \leftarrow \mathbf{u}_{k} +\frac{\sum_{m=1}^{M} \exp \Bigl( \frac{-1}{\lambda} \left[\tilde{S}\left(\tau_{m}\right) -\tilde{S}_{\min} \right] \Bigr) \delta \mathbf{u}_{k, m}}{\sum_{m=1}^{M} \exp \Bigl( \frac{-1}{\lambda} \left[\tilde{S}\left(\tau_{m}\right) -\tilde{S}_{\min} \right] \Bigr)},
 \end{equation}
which incorporates the \textit{cost-to-go} of the $m^{\text{th}}$ trajectory $\tilde{S}\left(\tau_{m}\right)$, the minimum cost trajectory among all simulated rollouts $\tilde{S}_{\min}$ that prevents numerical overflow or underflow, without affecting the optimality of the algorithm, and the inverse temperature $\lambda$ that governs the selectiveness of the weighted average of trajectories. The resulting sequence is then smoothed with a Savitzky-Galoy filter \cite{savitzky1964smoothing}, followed by applying the first control $\mathbf{u}_{0}$ to the system, while the remaining sequence is utilized for warm-starting the next optimization step.


\section{Unscented Optimal Control}\label{Unscented Optimal Control}
In this section, we present a comprehensive overview of the unscented transform, followed by the unscented guidance technique, which combines the unscented transform with standard stochastic optimal control theory to generate a robust \textit{open-loop} method for managing system uncertainties. 
\vspace*{-10pt}
\subsection{Unscented Transform}
The Unscented Transform (UT), proposed by Julier and Uhlmann \cite{julier1995new}, aims to create nonlinear filters without the need for linearization.
It approximates a probability distribution function (PDF) after it passes through a non-linear transformation using a set of sampled points, known as sigma points \cite{knudsen2018stochastic, xu2020online}.

Formally, given the mean $\Bar{\mathbf{x}}_k$ and covariance $\mathbf{\Sigma}_k$ of 
a Gaussian-distributed system state $\mathbf{x}_k$, with $\mathbf{x}_{k} \sim \mathcal{N}(\Bar{\mathbf{x}}_k, \mathbf{\Sigma}_k) \;\ihab{\in \mathbb{R}^{n_x}}$,
UT approximates the distribution over the next state, $\mathbf{x}_{k+1}$, by first introducing a set of sigma points $\big\{\mathcal{X}_{k}^{(i)}\big\}^{2n_x}_{i = 0} \in \mathbb{R}^{n_\sigma}$ around the mean $\Bar{\mathbf{x}}_k$ and the corresponding weights $\big\{ w^{(i)} \big\}^{2n_x}_{i=0} \in \mathbb{R}^{n_\sigma}$, where \ihab{the number of sigma points is given by $n_\sigma = 2n_x + 1$ and $n_x$ is the dimension of the state space.}
These sigma points are designed to capture the covariance of the distribution at time-step $k$ as follows
\begin{equation}\label{eq:ut-sigma-points}
\begin{aligned}
 \mathcal{X}_k^{(0)} &= \Bar{\mathbf{x}}_k,\\
 \mathcal{X}_k^{(i)} &= \Bar{\mathbf{x}}_k + \Big(\sqrt{(n_x + \lambda_\sigma)\mathbf{\Sigma}_k}\Big)_{\!i}, \; \forall i =\{1, \dots, n_x\},\\  
 \mathcal{X}_k^{(i)} &= \Bar{\mathbf{x}}_k - \Big(\! \sqrt{(n_x \!+ \! \lambda_\sigma\!)\mathbf{\Sigma}_k}\Big)_{\!i}, \;\forall i \!=\!\{n_x \!+ \!1, \dots, 2n_x\!\}, 
\end{aligned}
\end{equation}
where $\Big(\sqrt{(n_x + \lambda_\sigma)\mathbf{\Sigma}_k}\Big)_{i}$ represents the $i^{\text{th}}$ row or column of the square root of the weighted covariance matrix $(n_x + \lambda_\sigma)\mathbf{\Sigma}_k$,
and $\lambda_\sigma  = \alpha^2(n_x + k_\sigma) - n_x$ is influenced by the scaling parameters $k_\sigma \geq 0$ and $\alpha \in (0,1]$ that determine how far the sigma points are spread from the mean~\cite{van2004sigma}, as demonstrated in Fig.~\ref{fig:generated-samples}.
Each $\mathcal{X}_{k}^{(i)}$ is associated with two weights, $w^{(i)}_m$ for computing the mean and $w^{(i)}_c$ for determining the covariance of the transformed distribution, computed as
\begin{equation}\label{eq:ut-weights}
\begin{aligned}
 w_{m}^{(0)} &= \frac{\lambda_\sigma}{n_x + \lambda_\sigma},\\ 
 w_{c}^{(0)} &= w_{m}^{(0)} + (1 - \alpha^2 + \beta),\\  
 w_{m}^{(i)} &= w_{c}^{(i)} = \frac{1}{2(n_x + \lambda_\sigma)}, \junhong{\; \forall i =\{1, \dots, 2n_x\},}
\end{aligned}
\end{equation}
where $\beta$ is a hyper-parameter controlling the relative importance of the mean and covariance information.
In other words, $\beta$ is employed to incorporate prior knowledge about the distribution of state $\mathbf{x}$. For Gaussian distributions, the optimal value for $\beta$ is 2 \cite{julier2002scaled}.
In the second step, we propagate the $(2n_x + 1)$ sigma points through the non-linear system to produce the transformed sigma points $\mathcal{X}_{k+1}^{(i)} = f(\mathcal{X}_k^{(i)})$ at the next time-step. 
Finally, the mean and covariance of $\mathbf{x}_{k+1}$ can be estimated using the transformed sigma points and their corresponding weights, as follows
\begin{equation}\label{eq:ut-mean-cov}
\begin{aligned}
   \mathbf{\Bar{x}}_{k+1} &= \sum_{i=0}^{2n_x} w_{m}^{(i)}\mathcal{X}_{k+1}^{(i)}, \\
   \mathbf{\Sigma}_{k+1} &= \sum_{i=0}^{2n_x} w_{c}^{(i)}(\mathcal{X}_{k+1}^{(i)} - \mathbf{\Bar{x}}_{k+1})(\mathcal{X}_{k+1}^{(i)} - \mathbf{\Bar{x}}_{k+1})^{\top}.
\end{aligned}
\end{equation}
\subsection{Unscented Optimal Control}\label{Sub: Unscented Optimal Control}
By incorporating the unscented transform with standard optimal control, the unscented optimal control, also referred to as unscented guidance \cite{ross2015unscented}, presents a novel methodology for addressing the uncertainties in non-linear dynamical systems within an \textit{open-loop} framework \cite{ross2014unscented, ozaki2020tube}. 
Given the sigma points $\mathcal{X}_k^{(i)}$ and disturbed control input $\mathbf{w}_{k}$ at time-step $k$, each sigma point can be propagated through the underlying non-linear dynamics given in (\ref{eq: underlying non-linear dynamics}), as follows
\begin{equation}\label{eq:dynamics of sigma points}
\mathcal{X}_{k+1}^{(i)} = f\left(\mathcal{X}_k^{(i)}, \mathbf{w}_{k}\right), 
\forall i = 0, \cdots, 2n_x \in \mathbb{R}^{n_\sigma}.
\end{equation}
Consider an $n_\sigma n_x$-dimensional vector $\mathbf{X}$, defined as
$
\mathbf{X}=\left[\mathcal{X}^{(0)}, \mathcal{X}^{(1)}, \ldots, \mathcal{X}^{\left(2n_x\right)} \right]^{\top}  \in \mathbb{R}^{n_x n_\sigma} $. 
Then, the dynamics of $\mathbf{X}$ are characterized by $n_\sigma$ instances of the function $f$, specified as
\begin{equation}\label{eq:dynamics of ALL sigma points}
\boldsymbol{X}_{k+1}=\left[\begin{array}{c}
f\left(\mathcal{X}_k^{(0)}, \mathbf{w}_k\right) \\
f\left(\mathcal{X}_k^{(1)}, \mathbf{w}_k\right) \\
\vdots \\
f\left(\mathcal{X}_k^{(2n_x)}, \mathbf{w}_k\right)
\end{array}\right]:=\mathbf{f}(\boldsymbol{X}_k, \mathbf{w}_k).
\end{equation}
With these preliminaries, the original stochastic optimal control problem described in (\ref{eq:2}) can be re-formulated within the context of the unscented guidance framework as
\vspace*{-3pt}
\begin{subequations}
\begin{align}
\min _{\mathbf{U}} \; \!\mathbf{J} (\mathbf{X}&,\mathbf{u}) =  \mathbb{E}\!\left[\mathbf{\Phi 
 }\!\left(\mathbf{X}_{N}\!\right)\!+\!\!\!\sum_{k=0}^{N-1}\!\!\left(\!\mathbf{q}\!\left(\mathbf{X}_{k}\right)\!+\!\frac{1}{2} \mathbf{u}_{k}^{\top}  R \mathbf{u}_{k}\!\!\right)\!\!\right]\!\!, \label{eq:uoc_a}\\
\text {s.t.} 
\quad & \mathbf{X}_{k+1}=\mathbf{f}\left(\mathbf{X}_{k}, \mathbf{w}_{k}\right), \delta \mathbf{u}_{k} \sim \mathcal{N}(\mathbf{0}, \Sigma_{\mathbf{u}}), \label{eq:uoc_b}\\
& \mathcal{X}_{rob}\left(\mathbf{X}_{k}\right) \cap \mathcal{X}_{obs}=\emptyset, \;\mathbf{h}(\mathbf{X}_k, \mathbf{u}_k) \leq 0, \label{eq:uoc_c}\\
& \mathbf{X}_0 = \left[\mathcal{X}_0^{(0)}\!, \ldots, \mathcal{X}_0^{(2n_x)}\right]^{\top} \!\!, \mathbf{u}_{k} \!\in \mathbb{U},\mathbf{X}_{k} \!\in \mathbb{X},
\end{align}
\label{eq: unscented optimal control problem}
\end{subequations}
where $\mathbf{\Phi} \left(\mathbf{X}_{N} \right)\!\! = \!\!\left[\phi \left(\mathcal{X}_{N}^{(0)}\right), \ldots, \phi \left(\mathcal{X}_{N}^{(2n_x)}\right)\right]^{\top}  \!\!\! \in \! \mathbb{R}^{n_\sigma}$, 
and 
$\mathbf{q} \left(\mathbf{X}_k \right) = \left[q\left(\mathcal{X}^{(0)}_k \right), \ldots, q\left(\mathcal{X}^{(2n_x)}_k \right)\right]^{\top}  \!\! \in \mathbb{R}^{n_\sigma}$.
The objective of our proposed U-MPPI control strategy is to minimize the objective function, $\mathbf{J}$, in (\ref{eq:uoc_a}) by finding the optimal control sequence, $\mathbf{U} = \left\{\mathbf{u}_{k}\right\}_{k=0}^{N-1}$, while taking into account (i) the system 
constraints previously discussed in Section \ref{Problem Formulation}, and (ii) the uncertainties associated with both system states and control actions. 
In U-MPPI, minimizing the vector-valued cost function \( \mathbf{J} (\mathbf{X},\mathbf{u}) \), which represents the costs of sigma-point trajectories, is achieved through a sampling-based stochastic optimization approach similar to MPPI by: (i) leveraging the UT to propagate the sigma points, (ii) evaluating their trajectory costs using a risk-sensitive formulation, and (iii) updating the control sequence through a weighted sampling-based optimization, as detailed in the next section.
\begin{figure}[t!]
    \resizebox{1.\columnwidth}{!}
    {
    \includegraphics[scale=1]{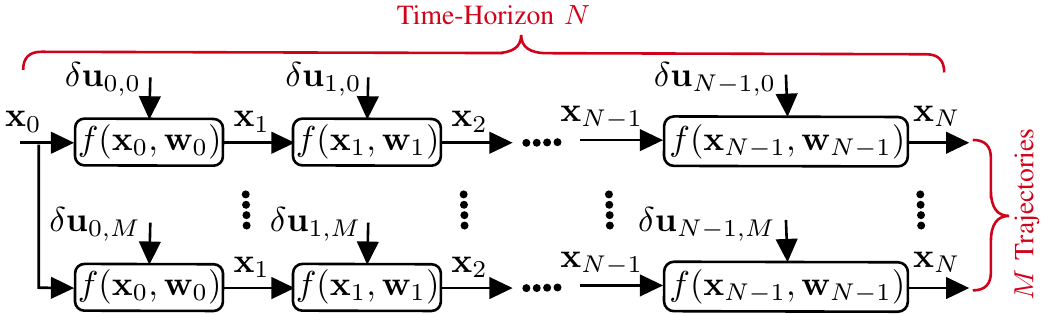}
    }
    \caption{Schematic illustration of system dynamics propagation in MPPI for $M$ sampled trajectories over a finite time-horizon $N$.
    }
    \label{fig:mppi_dynamical_propagation}%
\end{figure}

\section{U-MPPI Control Strategy}\label{U-MPPI Control Strategy}
As previously outlined in Section \ref{Overview of MPPI Control Strategy}, the control noise variance $\Sigma_{\mathbf{u}}$ is not updated by MPPI, and the state-space exploration is performed by adjusting $\nu$ (refer to ~\eqref{eq:cost-to-go}). Nevertheless, a too-high value of $\nu$ can cause control inputs with considerable chatter. 
Similarly, increasing $\Sigma_{\mathbf{u}}$ might violate system constraints and eventual divergence from the desired state 
\cite{mohamed2022autonomous}.
Additionally, the MPPI problem, as stated in \eqref{eq:2}, is focused solely on minimizing the cost function that is affected by a minor perturbation in the control input, represented by $\delta \mathbf{u}_{k}$, without explicitly incorporating the uncertainties that may be associated with either the system states or the surrounding environment. 
To ensure effectiveness in practice, the motion control strategy should be able to reflect the uncertainties of system states, sensing, and control actions. 
To this end, we introduce the U-MPPI control strategy, a new technique that leverages unscented transform to deal with these uncertainties.
More precisely, the UT is utilized for the purpose of regulating the propagation of the dynamical system, thereby introducing a novel and more efficient trajectory sampling strategy than the standard MPPI variants, as demonstrated in Section~\ref{Unscented-Based Sampling Strategy}. Furthermore, as discussed in Section~\ref{Risk-Sensitive Cost}, UT proposes a new cost function formulation incorporating uncertainty information, leading to a safer and more robust control system, especially for safety-critical applications.
\begin{figure*}[!ht]
    \resizebox{1.0\textwidth}{!}
    {
        \includegraphics[scale=1]{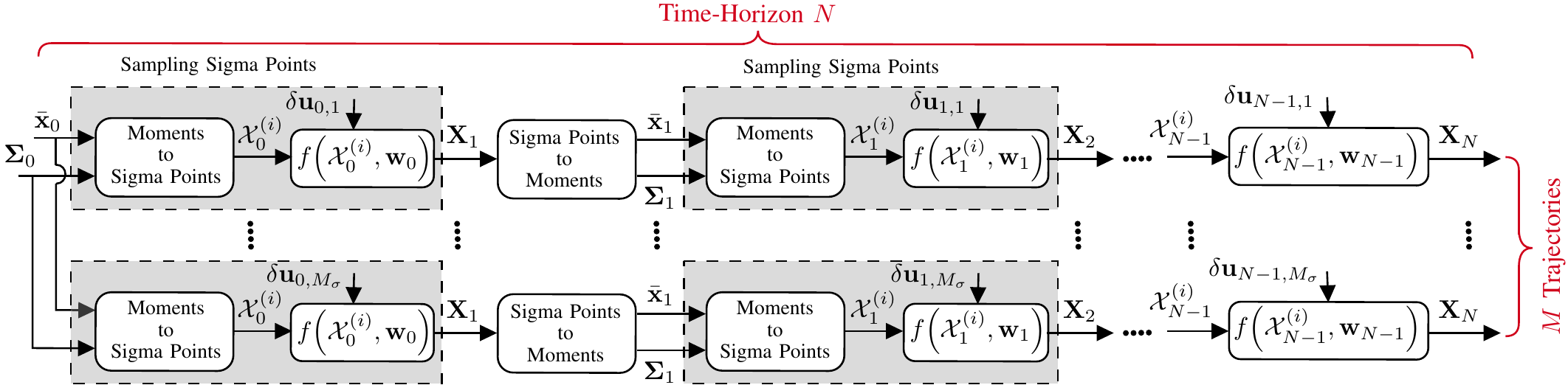}
    }
    \caption{Schematic illustration of nonlinear dynamical system propagation under the proposed U-MPPI control strategy for $M$ sampled trajectories over a finite time-horizon $N$, where $M=n_{\sigma} M_\sigma$.
    }%
    \label{fig:umppi_dynamical_propagation}%
\end{figure*}

\subsection{Unscented-Based Sampling Strategy}\label{Unscented-Based Sampling Strategy}
\ihab{\sout{By leveraging Monte Carlo simulation, the vanilla MPPI algorithm simulates a large number of \textit{real-time} trajectories $M$, propagated from the system dynamics defined in \eqref{eq: underlying non-linear dynamics}, by solely manipulating the injected Gaussian noise into the mean control sequence (see Fig.~\ref{fig:mppi_dynamical_propagation}). 
Additionally, the computation of the propagated states during the time period of $N$ is restricted to the mean value or the first moment with respect to the initial state $\mathbf{x}_0$, without propagating the covariance of $\mathbf{x}_k$.}} 
\ihab{
To gain a comprehensive understanding of our proposed unscented-based sampling strategy, it is essential to first examine how the system dynamics are propagated within the vanilla MPPI control strategy.
As illustrated in Fig.~\ref{fig:mppi_dynamical_propagation}, given the current robot state \( \mathbf{x}_0 \) and an \( N \times M \) matrix of random control variations \( \delta \mathbf{u} \) generated on a GPU using CUDA’s random number generation library, the MPPI algorithm optimizes control inputs by propagating the system dynamics over a specified time-horizon \( N \) for \( M \) trajectories. 
To be more precise, for a particular trajectory \( m \in \{1, \cdots, M\} \) (represented by each row), the process involves generating a control input sequence over the time horizon \( N \) by perturbing the nominal inputs \( \left\{\mathbf{u}_{k}\right\}_{k=0}^{N-1} \), which represent the optimal control sequence provided by MPPI during the previous control loop interval \(\Delta t\) as previously explained in Section~\ref{Overview of MPPI Control Strategy}, with \( \left\{\delta \mathbf{u}_{k, m}\right\}_{k=0}^{N-1} \). At each discrete time-step \( k \), the next state \( \mathbf{x}_{k+1} \) is computed using the system’s dynamic model \( f \) as defined in \eqref{eq: underlying non-linear dynamics}, such that \( \mathbf{x}_{k+1} = f(\mathbf{x}_k, \mathbf{u}_k + \delta \mathbf{u}_{k,m}) \). This results in a sequence of states \( \{\mathbf{x}_1, \mathbf{x}_2, \ldots, \mathbf{x}_N\} \equiv \big\{\mathbf{x}_k\big\}^{N}_{k = 1} \). This process is repeated for multiple trajectories by varying the perturbations \( \left\{\delta \mathbf{u}_{k, m}\right\}_{k=0}^{N-1} \), resulting in \( M \) trajectories. Each trajectory is then evaluated using a predefined cost function, incorporating factors such as target deviation and control effort.
It is important to notice that, as illustrated in the schematic diagram, the vanilla MPPI algorithm leverages Monte Carlo simulation to simulate a large number of real-time trajectories \( M \) by solely manipulating the injected Gaussian noise \( \delta \mathbf{u} \) into the mean control sequence \( \mathbf{U} \). Moreover, within the time horizon \( N \), only the first moment (i.e., mean) of the propagated states is computed given the initial state \( \mathbf{x}_0 \), without propagating the covariance of \( \mathbf{x}_k \). To enhance the performance of the classic MPPI algorithm, we propose a new trajectory sampling technique that utilizes the UT to propagate both the mean \( \bar{\mathbf{x}}_k \) and covariance \( \mathbf{\Sigma}_k \) of the state vector \( \mathbf{x}_k \) at each discrete time-step \( k \).
}

\ihab{}
\ihab{
In the proposed U-MPPI control strategy, as displayed in Fig.~\ref{fig:umppi_dynamical_propagation}, given the current robot state $\mathbf{x}_{0} \sim \mathcal{N}(\Bar{\mathbf{x}}_0, \mathbf{\Sigma}_0)$ and $ N \times M_\sigma$ random control perturbations, the system dynamics are propagated over the entire time-horizon \( N \) by converting the moments of the state distribution into sigma points at each time-step \( k \), then propagating these sigma points through the system's dynamic model \( f \) as expressed in \eqref{eq:dynamics of sigma points}, followed by converting the propagated sigma points back into moments to compute the updated mean state and covariance at \( k+1 \). 
This methodology leads to a total of \( M \) rollouts, achieved by sampling \( M_\sigma \) sets of batches, also referred to as \textit{cones}, with each batch comprising \( n_\sigma \) trajectories, such that \( M = n_\sigma M_\sigma \) (each row in the figure represents a batch).
For each batch \( m \in \{1, \cdots, M_\sigma\} \), the propagation process of our proposed sampling strategy involves generating a control input sequence over the time-horizon \( N \) by perturbing the nominal inputs \( \left\{\mathbf{u}_{k}\right\}_{k=0}^{N-1} \) with \( \left\{\delta \mathbf{u}_{k,m}\right\}_{k=0}^{N-1} \), similar to the MPPI propagation process. First, the initial mean state \( \mathbf{\Bar{x}}_0 \) and its covariance \( \mathbf{\Sigma}_0 \) are used to generate a set of sigma points \( \big\{\mathcal{X}_{0}^{(i)}\big\}^{2n_x}_{i = 0} \) via \eqref{eq:ut-sigma-points}. 
Afterwards, at each subsequent time-step \( k \), these sigma points \( \big\{\mathcal{X}_{k}^{(i)}\big\}^{2n_x}_{i = 0} \) are propagated through the nonlinear dynamics expressed in \eqref{eq:dynamics of sigma points} as \(\mathcal{X}_{k+1}^{(i)} = f\left(\mathcal{X}_k^{(i)}, \mathbf{u}_k + \delta \mathbf{u}_{k,m}\right)\), along with the control inputs perturbed by Gaussian noise \( \delta \mathbf{u}_{k,m} \). This results in a new set of propagated sigma points \( \big\{\mathcal{X}_{k+1}^{(i)}\big\}^{2n_x}_{i = 0} \), which can be merged into a single vector \( \boldsymbol{X}_{k+1} \) using \eqref{eq:dynamics of ALL sigma points}.
The propagated sigma points are then converted back into moments by applying \eqref{eq:ut-mean-cov} to compute the updated mean state \( \Bar{\mathbf{x}}_{k+1} \) and covariance \( \mathbf{\Sigma}_{k+1} \). This iterative process of ``\textit{Moments to Sigma Points}'' and ``\textit{Sigma Points to Moments}'' is repeated for each time-step \( k \) over the entire time-horizon \( N \), ensuring accurate propagation of both the mean and covariance of the state distribution. This results in a sequence of state vectors \sout{\( \big\{\mathbf{X}_k\big\}^{N}_{i = 1} \)} \( \{\mathbf{X}_1, \mathbf{X}_2, \ldots, \mathbf{X}_N\} \equiv \big\{\mathbf{X}_k\big\}^{N}_{k = 1} \), representing the \( n_\sigma \) propagated sigma points. 
This sequence generation is performed for multiple trajectories \( M_\sigma \) by varying the perturbations \( \delta \mathbf{u}_{k,m} \) for each trajectory \( m \), resulting in a set of \( M \) possible state trajectories. 
In this study, we assume that during each time-step \( k \), all sigma points belonging to the \( m^{\text{th}} \) batch are influenced by the same Gaussian control noise, represented by \( \delta \mathbf{u}_{k,m} \).
Figure~\ref{fig:vis-ut-sampling-strategy} provides a visual representation of our proposed unscented-based sampling strategy depicted in Fig.~\ref{fig:umppi_dynamical_propagation}.
}

\begin{figure}[t]%
\vspace*{-10pt}
    \centering
    \subfloat[$\delta \mathbf{u}_{k} \sim \mathcal{N}(\mathbf{0},0.025\mathbf{I}_2)$]{\vspace*{-1pt}{\includegraphics[scale=0.95]{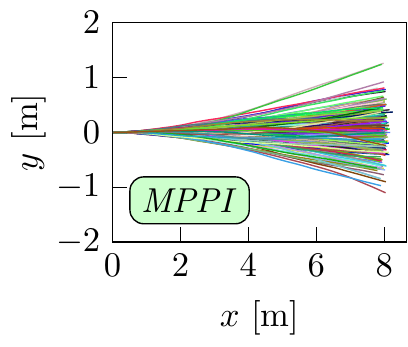}}\label{fig:samples-normal}}%
    \quad
   \subfloat[$\alpha=1, k_\sigma = 0, \mathbf{\Sigma}_0 = 0.01\mathbf{I}_3$]{\vspace*{0pt}{\includegraphics[scale=.95]{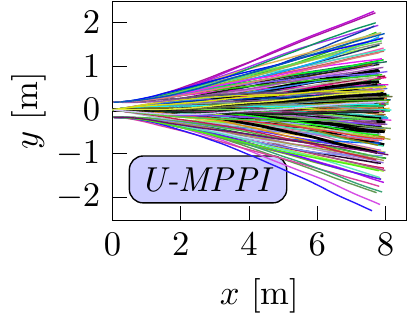}}\label{fig:samples-ut_k_0_alpha_1_sigma_.01}}\\
    \subfloat[\!$\alpha=1\!, k_\sigma = 3, \mathbf{\Sigma}_0 = 0.012\mathbf{I}_3$]{\vspace*{0pt}{\includegraphics[scale=0.95]{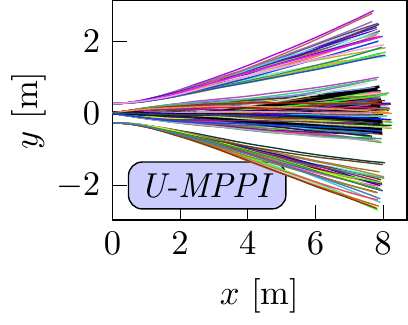}}\label{fig:samples-ut_k_3_alpha_1_sigma_.012}}%
    \quad
    \subfloat[\!$\alpha \!=0.1, k_\sigma \!= 0, \mathbf{\Sigma}_0 \! = 0.01\mathbf{I}_3$]{\vspace*{0pt}{\includegraphics[scale=.95]{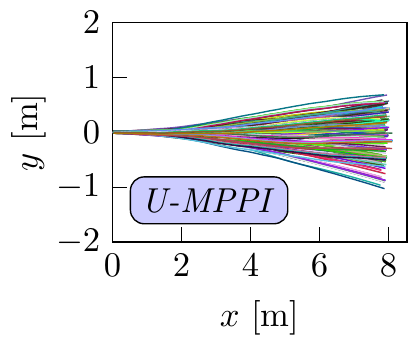}}\label{fig:samples-ut_k_0_alpha_0.1_sigma_.01}}%
    \caption{Distribution of $210$ sampled trajectories generated by (a) MPPI with $\delta \mathbf{u}_{k} \sim \mathcal{N}(\mathbf{0}, 0.025\mathbf{I}_2)$ and (b) U-MPPI with the same perturbation in the control input $\delta \mathbf{u}_{k}$ but with different UT parameters; in both methods, the robot is assumed to be initially located at $\mathbf{x} = [{x}, {y}, \theta]^{\top}= [0,0,0]^{\top}$ in ([\si{\metre}], [\si{\metre}], [\si{\deg}]), with a commanded control input $\mathbf{u} = [v,\omega]^{\top} = [1, 0]^{\top}$ in ([\si{\metre/\second}], [\si{
\radian/\second}]).}%
    \label{fig:generated-samples}%
\end{figure}

The primary objective of the new sampling strategy is to achieve significantly better exploration of the state-space of the given system and more efficient sampling of trajectories compared to MPPI, while utilizing the same injected control noise $\Sigma_{\mathbf{u}}$.
To exemplify how leveraging the UT for trajectory sampling can enhance the performance of the MPPI algorithm and to investigate the effect of UT parameters on the distribution of sampled rollouts, we present a concrete example in Fig.~\ref{fig:generated-samples}.
In particular, \ihab{using the discrete-time kinematics model of a differential wheeled robot described in \eqref{eq:differential-drive-robot}} and control schemes parameters listed in Section~\ref{Simulation Setup:Cluttered Environments}, we generate \num{210} rollouts by sampling $\delta \mathbf{u}_{k}$ from a zero-mean Gaussian distribution with a covariance of $0.025\mathbf{I}_2$ under the classical MPPI framework, as illustrated in Fig.~\ref{fig:samples-normal}, where $\mathbf{I}_n$ denotes an $n \times n$ identity matrix.
Similarly, we employ the unscented-based sampling strategy to draw \num{210} trajectories, considering: (i) the same injected disturbance into the control input, i.e., $\delta \mathbf{u}_{k} \sim \mathcal{N}(\mathbf{0},0.025\mathbf{I}_2)$, and (ii) setting the UT scaling parameters to $\alpha=1$ and $k_\sigma = 0$, along with an initial state covariance matrix $\mathbf{\Sigma}_0$ of $0.01\mathbf{I}_3$, as illustrated in Fig.~\ref{fig:samples-ut_k_0_alpha_1_sigma_.01}.
It is noteworthy to observe in Fig.~\ref{fig:samples-ut_k_0_alpha_1_sigma_.01} that our proposed sampling strategy is more efficient than the classical MPPI sampling strategy, as it generates more spread-out trajectories that cover a larger state space. This enables the robot to explore the environment more extensively and find better solutions, thereby reducing the likelihood of getting trapped in local minima, as revealed in Section~\ref{Simulation Results: 1}.  
Nevertheless, as depicted in Fig.~\ref{fig:samples-ut_k_3_alpha_1_sigma_.012}, employing higher values of UT parameters (specifically, $\alpha, k_\sigma, \mathbf{\Sigma}_0$) may result in a loss of precision and continuity in the distribution of trajectories across the state-space, as the sigma points become more spread out from the mean $\mathcal{X}_k^{(0)}$. This can impact the resulting control actions of the system and, in the context of autonomous navigation, potentially lead to collisions with obstacles. In contrast, Fig.~\ref{fig:samples-ut_k_0_alpha_0.1_sigma_.01} demonstrates that using lower values of $\alpha$ and $k_\sigma$ results in a sampling strategy that closely resembles MPPI. 
Similarly, if trajectories are sampled only from $\mathcal{X}_k^{(0)}$ (as depicted by the blue trajectories in Fig.~\ref{fig:vis-ut-sampling-strategy}), while excluding other sigma-point trajectories, the same sampling strategy can be achieved. 
We refer to this approach as sampling mode 0 ($\texttt{SM}_0$), while the default strategy that includes all sigma points is referred to as $\texttt{SM}_1$.

\subsection{Risk-Sensitive Cost}\label{Risk-Sensitive Cost}
One of the main limitations of the vanilla MPPI is that it typically assumes a \textit{risk-neutral} approach when assessing the sampled trajectories during the optimization process, without explicitly considering risk or uncertainty in
the trajectory evaluation process, as outlined in Section~\ref{Overview of MPPI Control Strategy}, particularly in ~\eqref{eq:cost-to-go}. 
The commonly employed method in sampling-based MPC algorithms 
involves using a quadratic cost function to guide the current state $\mathbf{x}_k$ towards its desired state $\mathbf{x}_f$, denoted as $q_{\text{state}}(\mathbf{x}_k)$, and expressed mathematically as follows
\begin{equation}\label{eq:quadratic-state}
  \begin{aligned}
q_{\text{state}}(\mathbf{x}_{k}) =\left(\mathbf{x}_k -\mathbf{x}_{f}\right)^{\top} \mathrm{Q}\left(\mathbf{x}_{k}-\mathbf{x}_{f}\right) =\left\|\mathbf{x}_{k}-\mathbf{x}_{f}\right\|_{\mathrm{Q}}^2,
\end{aligned}  
\end{equation}
where $\mathrm{Q}$ is a positive definite weighting matrix. \ihab{Hence, our key objective in this work is to integrate a risk measure into the U-MPPI optimization problem, enabling the adjustment of responses to varying uncertainty levels and improving adaptability over the conventional MPPI algorithm.}

Whittle introduced in \cite{whittle1981risk} an interesting \ihab{risk measure} method for incorporating risk in decision-making by replacing the expected quadratic cost with a risk-sensitive benchmark in the form of an exponential-quadratic function. 
\ihab{Such a formulation maps uncertainties to a numerical value to assess the potential threat from extreme events. In the context of risk-sensitive control, this involves assessing the impact of uncertainty on system performance and adjusting responses to manage risks effectively.
Mathematically,} the \textit{risk-sensitive} (RS) cost $q_{\mathrm{rs}}\left(\mathbf{x}_k\right)$ can be obtained by evaluating the log-expectation of the exponentiated quadratic cost as follows
\begin{equation}\label{Q_rs_1st-form}
\begin{aligned}
q_{\mathrm{rs}}(\mathbf{x}_k) &= -\frac{2}{\gamma} \log \mathbb{E}\left[\exp\left(-\frac{1}{2} \gamma q_{\text{state}}(\mathbf{x}_{k}) \right)\right] \\
&= -\frac{2}{\gamma} \log \mathbb{E}\left[\exp\left(-\frac{1}{2} \gamma 
\left\|\mathbf{x}_{k}-\mathbf{x}_{f}\right\|_{\mathrm{Q}}^2\right)\right].
\end{aligned}
\end{equation}
\ihab{Notably, as \(\gamma\) approaches zero, the RS cost \( q_{\mathrm{rs}}(\mathbf{x}_k) \) converges to the standard quadratic cost \( q_{\text{state}}(\mathbf{x}_k) \),
i.e.,\footnote{\ihab{This is because, as \(\gamma\) becomes very small, the exponential function can be approximated using a Taylor series expansion, and the logarithmic term can be simplified accordingly.}} 
\[\lim_{\gamma \to 0} q_{\mathrm{rs}}(\mathbf{x}_k) = \left\|\mathbf{x}_{k}-\mathbf{x}_{f}\right\|_{\mathrm{Q}}^2 = q_{\text{state}}(\mathbf{x}_k),
\] 
thus avoiding the singularity issue when \(\gamma = 0\). }
Here, $\gamma$ is a real scalar denoted as the \textit{risk-sensitivity} parameter, dictating how the controller reacts to risk or uncertainty.
For example, when $\gamma > 0$, the controller exhibits \textit{risk-seeking} or \textit{risk-preferring} behavior\ihab{,
favoring trajectories that may offer higher rewards, even if they come with higher uncertainties or risks. 
As a result, a controller minimizing the \textit{risk-seeking} cost \( q_{\mathrm{rs}} \) will actively steer the system state \( \mathbf{x}_{k} \) towards areas of uncertainty (i.e., uncertain regions) where the cost \( q_{\mathrm{rs}} \) imposes lower penalties. Such behavior encourages the exploration of unknown or uncertain areas, which can lead to a more comprehensive understanding of the system dynamics.}
Conversely, when $\gamma < 0$, the controller demonstrates \textit{risk-averse} or \textit{risk-avoiding} behavior,
\ihab{prioritizing safer trajectories and leading to more conservative control actions. In this scenario, the controller minimizes \textit{risk-averse} cost \( q_{\mathrm{rs}} \) by driving the system state towards areas with minimal uncertainty and lower penalties. This approach prioritizes stability and safety by avoiding risky or uncertain areas, making it suitable for applications where minimizing risk is crucial. Examples include autonomous medical robots performing delicate surgeries, where precision and safety are crucial to prevent patient harm.
It is noteworthy that a control strategy aimed at minimizing the maximum possible loss (i.e., solving a minimax optimization problem) aligns with the \textit{risk-averse} behavior inherent in the risk-sensitive function, thereby representing a specific instance of risk-sensitive control designed to ensure robust and reliable performance. As discussed in \cite{lofberg2003minimax} (see Chapter 8), the author examines the relationship between \textit{risk-sensitive} and \textit{minimax} optimization problems in MPC, allowing us to assert that a controller designed to minimize the worst-case scenario (i.e., a minimax control approach) can be interpreted as minimizing a \textit{risk-averse} performance measure (i.e., objective function).
} 
 When $\gamma = 0$, the controller is considered \textit{risk-neutral}\ihab{, treating all uncertainties uniformly without preference towards risk. 
 The risk-sensitivity parameter \(\gamma\) thus directly influences the optimization problem described in \eqref{eq: unscented optimal control problem}, 
determining whether the control policy will adopt \textit{risk-seeking} or \textit{risk-averse} behaviors based on the selected value of \(\gamma\). 
 }
 We refer to \cite{yang2015risk} for more details and clear visualizations demonstrating the relationship between the sign of $\gamma$ and the RS cost $q_{\mathrm{rs}}$.

\begin{figure}[t]
    \centering
    \includegraphics[scale=1]{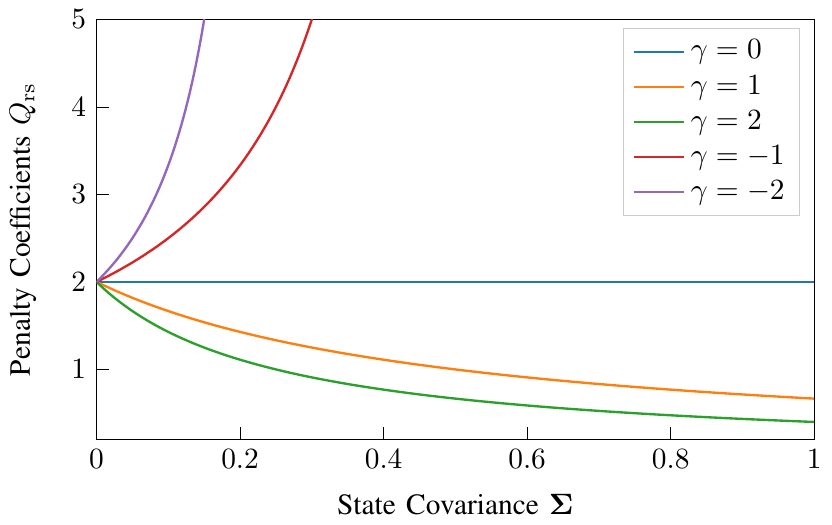}
    \caption{\ihab{Influence of different \(\gamma\) values on the penalty coefficients \( Q_{\mathrm{rs}} \) for state reference tracking with \(Q = 2\) and varying 1-dimensional state covariance \(\mathbf{\Sigma}\).}}
    \label{fig:risk-cost-plot}
\end{figure}

\ihab{Building on Whittle's foundational work, our proposed U-MPPI control strategy leverages this risk-sensitive approach to enhance the robustness and safety of autonomous vehicle navigation. By integrating the exponential-quadratic cost function \( q_{\mathrm{rs}} \) into our trajectory evaluation process, we can effectively manage system uncertainties and enhance overall control efficiency.}
Our control strategy assumes that the system state $\mathbf{x}_{k} \sim \mathcal{N}(\mathbf{\Bar{\mathbf{x}}}_k, \mathbf{\Sigma}_k)$ follows Gaussian distribution. With this assumption, we can approximate the RS state-dependent cost expressed in \eqref{Q_rs_1st-form} as
\begin{equation}\label{Qrs}
q_{\mathrm{rs}}(\mathbf{x}_{k})
=
\frac{1}{\gamma} \log \operatorname{det}\left(\mathbf{I}+\gamma Q \mathbf{\Sigma}_{k}\right)
+\left\|\mathbf{\Bar{\mathbf{x}}}_{k}-\mathbf{x}_{f}\right\|_{Q_{\mathrm{rs}}}^2,
\vspace*{-3pt}
\end{equation}
where $Q_{\mathrm{rs}}$ represents the \textit{risk-sensitive} penalty coefficients matrix or adaptive weighting matrix, given by $Q_{\mathrm{rs}}(\mathbf{\Sigma}_{k}) = {\left(Q^{-1}+\gamma \mathbf{\Sigma}_{k}\right)^{-1}}$. 
The derivation of \eqref{Qrs} is given in Appendix \ref{Appendix Risk Sensitive Cost Function}.
Such a new formulation is employed by U-MPPI to assess each batch of sigma-point trajectories, where the predicted mean state $\Bar{\mathbf{x}}_{k}$ is replaced with the predicted sigma points $\big\{\mathcal{X}_{k}^{(i)}\big\}^{2n_x}_{i = 0}$. Therefore, the modified RS cost for the $i^{\text{th}}$ U-MPPI sampled trajectory within a certain batch is defined by
\begin{equation}\label{Qrs-modified}
q_{\mathrm{rs}}\!\left(\!\mathcal{X}_{k}^{(i)}\!\!, \mathbf{\Sigma}_{k}\!\right) =
\frac{1}{\gamma} \log \operatorname{det}\left(\mathbf{I}\!\,+\!\,\gamma Q \mathbf{\Sigma}_{k}\right)
+\left\|\mathcal{X}_{k}^{(i)}\!\!-\mathbf{x}_{f}\right\|_{Q_{\mathrm{rs}}}^2\!\!\!\!.
\vspace*{-2pt}
\end{equation}
It is noteworthy to observe in \eqref{Qrs-modified} that $q_{\mathrm{rs}}(\cdot)$ incorporates uncertainty information $\mathbf{\Sigma}_{k}$ as a feedback into the weighting matrix $Q$, which measures the difference between the predicted sigma point $\mathcal{X}_{k}^{(i)}$ and the desired state $\mathbf{x}_{f}$. 
By incorporating this uncertainty feedback mechanism, the proposed control strategy exhibits a risk-sensitive behavior that effectively minimizes the RS cost over the time-horizon $N$. 
\ihab{Additionally, by utilizing the relationship between \(\gamma\) and the penalty on state reference tracking $Q_{\mathrm{rs}}$, we can adjust the behavior of our control strategy toward system uncertainty.
}
This, in turn, enables the development of a more robust and risk-conscious control policy, as empirically validated through intensive simulations outlined in Section \ref{Simulation Details and Results}. 
To be more precise, if $\gamma<0$, the penalty coefficients matrix $Q_{\mathrm{rs}}$ utilized for tracking the desired state increases with the level of system uncertainty $\mathbf{\Sigma}_{k}$. This can be expressed mathematically over a finite time-horizon $N$ as 
\[ Q_{\mathrm{rs}}(\mathbf{\Sigma}_{0}) < Q_{\mathrm{rs}}(\mathbf{\Sigma}_{1}) < \dots < Q_{\mathrm{rs}}(\mathbf{\Sigma}_{N-1}) \quad \forall \gamma < 0. \] 
\ihab{Consequently, the RS cost function $q_{\mathrm{rs}}$ also increases, reflecting the higher penalties imposed for deviation from the desired state $\mathbf{x}_{f}$. In the context of autonomous navigation in a cluttered environment, this means that as uncertainty increases, the control strategy becomes more aggressive in trying to reach the desired state, leading to higher chances of collisions and aggressive maneuvers.}
On the other hand, if $\gamma>0$, the penalty matrix decreases as the uncertainty level increases. \ihab{As a result, the cost function \(q_{\mathrm{rs}}\) decreases, indicating lower penalties for deviation from the desired state under higher uncertainty. This results in a more conservative approach, where the control policy is more exploratory and less forceful in reaching the desired state, thereby reducing the risk of collisions.} Additionally, when $\gamma = 0$, the penalty remains constant and equal to the weighting matrix $Q$, regardless of the level of the uncertainty. 
 \ihab{
%
Figure~\ref{fig:risk-cost-plot} provides, as an example, an intuitive visualization of how the risk-sensitivity parameter \(\gamma\) impacts the penalty coefficients \( Q_{\mathrm{rs}} \) for state reference tracking, demonstrating transitions between \textit{risk-seeking} (\(\gamma > 0\)) and \textit{risk-averse} (\(\gamma < 0\)) behaviors across various sizes of state covariance \(\mathbf{\Sigma}\).\footnote{\ihab{It is noteworthy that when \(\gamma < 0\),  \( Q_{\mathrm{rs}} \) may not always be positive. To ensure \( Q_{\mathrm{rs}} \) remains positive in this scenario, the condition \(\mathbf{\Sigma} < -\frac{1}{\gamma Q}\) must be met, where \( Q \) and \(\mathbf{\Sigma}\) are scalars. For an \( n \)-dimensional \(\mathbf{\Sigma}\) and \( Q \), ensuring a positive definite \( Q_{\mathrm{rs}} \) when \(\gamma < 0\) is discussed in \cite{hyeon2020fast}, which presents one possible solution by bounding \( Q_{\mathrm{rs}} \) using a sigmoid function.
}} For instance, in the case of \(\gamma = 1\), the feedback provided by a large system uncertainty \(\mathbf{\Sigma}\) will significantly reduce the penalty coefficients \( Q_{\mathrm{rs}} \), thereby decreasing the cost \( q_{\mathrm{rs}} \) \cite{hyeon2020fast}.
Additionally, we can observe that assigning higher positive values to \(\gamma\) (e.g., increasing from \(\gamma = 1\) to \(\gamma = 2\)) results in \(Q_{\mathrm{rs}}\) decreasing more rapidly with increasing uncertainty \(\mathbf{\Sigma}\), signifying greater sensitivity to uncertainty. This leads to a more conservative control strategy, reducing penalties on state deviations.
In contrast, assigning higher negative values to \(\gamma\) (e.g., decreasing from \(\gamma = -1\) to \(\gamma = -2\)) causes \(Q_{\mathrm{rs}}\) to increase more sharply with rising uncertainty, indicating lower sensitivity to uncertainty. This fosters a more aggressive control approach, potentially resulting in higher penalties on state deviations.
Therefore, higher positive \(\gamma\) values enhance stability by lowering the penalty weight in uncertain conditions, whereas higher negative \(\gamma\) values may lead to instability by increasing the penalty weight.
}
More in-depth analysis of how the U-MPPI performance is influenced by the sign of $\gamma$ is discussed in Section~\ref{Simulation Setup:Cluttered Environments}.

By incorporating the system uncertainty $\mathbf{\Sigma}_{k}$ into \eqref{cost-to-go-mppi}, we can obtain the modified \textit{cost-to-go} for each trajectory $\tau^{(i)}_m$ in batch $m$ as 
\vspace*{-5pt}
\begin{equation}\label{eq:cost-to-go-umppi}
\begin{aligned}
 \tilde{S}\left(\tau^{(i)}_m \right) = &\,\phi\left(\mathcal{X}_{N}^{(i)} \right) + \sum_{k=0}^{N-1} \tilde{q}\left(\mathcal{X}_{k}^{(i)}, \mathbf{\Sigma}_{k}, \mathbf{u}_{k}, \delta \mathbf{u}_{k,m}\right), \\ 
 &\forall m \in  \{1, \cdots, M_\sigma\}, \quad \forall i \in  \{0, \cdots, 2n_x\},
 \end{aligned}
\end{equation}
where the instantaneous running cost $\tilde{q}$ is a combination of the state-dependent running cost $q\left(\mathcal{X}_{k}^{(i)}, \mathbf{\Sigma}_{k}\right)$, which relies on the RS cost defined in \eqref{Qrs-modified} (as shown in \eqref{eq:state-dep-cost-function-umppi} as an example), as well as the quadratic control cost \ihab{\sout{$q\left(\mathbf{u}_{k}, \delta \mathbf{u}_{k} \right)$} $q_{\mathbf{u}}\left(\mathbf{u}_{k}, \delta \mathbf{u}_{k} \right)$} from \eqref{eq:cost-to-go}. Note that the \textit{cost-to-go} for all sigma-point trajectories in the $m^{\text{th}}$ batch can be expressed in vector form as
$\mathbf{\tilde{S}}\left(\Tau_{m}\right) = \left[\tilde{S}\left(\tau^{(0)}_{m}\right), \dots, \tilde{S}\left(\tau^{(2n_x)}_{m}\right)\right]^\top \in \mathbb{R}^{n_\sigma}$. 
Similarly, $\mathbf{\tilde{q}}\left(\mathbf{X}_{k}, \mathbf{\Sigma}_k, \mathbf{u}_{k}, \delta \mathbf{u}_{k,m}\right) = \left[\tilde{q}\left(\mathcal{X}_{k}^{(0)},\cdot\right), \dots, \tilde{q}\left(\mathcal{X}_{k}^{(2n_x)},\cdot\right)\right]^\top$.


\begin{algorithm}[ht!]
\caption{\textit{Real-Time} U-MPPI Control Algorithm}
\label{alg:UMPPI-Alg.}
\hspace*{\algorithmicindent} \textbf{Given:} \\
\hspace*{1cm} $M, M_\sigma , N$: $\#$ of trajectories, batches, time-horizon, \\
\hspace*{1cm} $ \mathbf{f}, n_x, \Delta t$: Dynamics, state  dimension, time-step size,\\
\hspace*{1cm} $\phi, q, Q, \gamma, \lambda, \nu, \Sigma_{\mathbf{u}}, R$: Cost/Control parameters, \\
\hspace*{1cm} $\lambda_\sigma, k_\sigma, n_\sigma, \alpha, \beta, \mathbf{\Sigma}_0$: UT parameters, \\
\hspace*{1cm} SGF: Savitzky-Galoy (SG) convolutional filter, \\
\hspace*{\algorithmicindent} \textbf{Input:} \\
\hspace*{1cm} \ihab{$\mathbf{U} =$ \textit{ControlSequenceInitializer}($\mathbf{x}_\text{init}$),}\\
\hspace*{1cm} $\texttt{SM}$: Set U-MPPI sampling mode ($\texttt{SM}_0$ \textit{OR} $\texttt{SM}_1$),
\begin{algorithmic}[1] 
\While {\textit{task not completed}}
    \State $\Bar{\mathbf{x}}_{0} \leftarrow$ \textit{StateEstimator()}, \textcolor{blue}{\Comment{ $\Bar{\mathbf{x}}_{0} \in \mathbb{R}^{n_x}$}}
    \State $ \delta \mathbf{u} \leftarrow$ \textit{RandomNoiseGenerator$(\mathbf{0}, \Sigma_{\mathbf{u}}\!)$}, \textcolor{blue}{\Comment{\!$\delta \mathbf{u} \!\in \!\mathbb{R}^{N\!\, \times M_\sigma}$}}
    \For{$m \leftarrow 1$ \textbf{to} $M_\sigma$ \textit{in parallel}}
        \State $\left(\Bar{\mathbf{x}}, \mathbf{\Sigma} \right) \!\leftarrow \! \left(\Bar{\mathbf{x}}_0, \mathbf{\Sigma}_0 \right)\!,$ \textcolor{blue}{\Comment{\textit{Actual state} $\mathbf{x}_{0} \!\sim\! \mathcal{N}\!(\!\Bar{\mathbf{x}}_0, \mathbf{\Sigma}_0\!)$}}
        \State $\mathbf{\tilde{S}}\left(\Tau_{m}\right)  \leftarrow [0, \dots, 0]^\top,$ 
        \textcolor{blue}{\Comment{$\mathbf{\tilde{S}}\left(\Tau_{m}\right) \in \mathbb{R}^{n_\sigma}$}}
        \For{$k \leftarrow 0$ \textbf{to} $N-1$}
            \State $\mathbf{X}_{k} \leftarrow$ \textit{Moments2SigmaPoints$(\Bar{\mathbf{x}}_{k}, \mathbf{\Sigma}_k )$},
            \State $\mathbf{X}_{k+1} \leftarrow \mathbf{X}_{k}+ \mathbf{f}\left(\mathbf{X}_{k}, \ihab{g\left(\mathbf{u}_{k}+\delta \mathbf{u}_{k,m}\right)}\right) \Delta t$,
            \State $\mathbf{\tilde{S}}\left(\Tau_{m}\right) \!\leftarrow \!\mathbf{\tilde{S}}\left(\Tau_{m}\right)
            +
            \mathbf{\tilde{q}}\left(\mathbf{X}_{k}, \mathbf{\Sigma}_k, \mathbf{u}_{k}, \delta \mathbf{u}_{k,m}\right)$,
            \State $\left(\!\Bar{\mathbf{x}}_{k+1}\!\,, \mathbf{\Sigma}_{k+1}\!\right) \!\leftarrow$\! \textit{SigmaPoints2Moments$(\mathbf{X}_{k+1}\!)$},
        \EndFor
        \State $\mathbf{\tilde{S}}\left(\Tau_{m}\right) \leftarrow \mathbf{\tilde{S}}\left(\Tau_{m}\right)+ \mathbf{\Phi}\left(\mathbf{X}_{N}\right)$, 
    \EndFor
    \State $\tilde{S}_{\min} \leftarrow \min _{m}[\mathbf{\tilde{S}}\left(\Tau_{m}\right)]$, \hfill $\forall m=\{1, \dots, M_\sigma\}$
    \For{$k \leftarrow 0$ \textbf{to} $N-1$}
        \State $\mathbf{u}_{k} \!\leftarrow \! \textit{SGF}\!\left(\!\!\mathbf{u}_{k}
        \!+\!
        \frac{\sum_{m=1}^{M_\sigma} \exp \bigl( \!\frac{-1}{\lambda} \!\left[\mathbf{\tilde{S}}\left(\Tau_{m}\right) -\tilde{S}_{\min} \right] \bigr) \delta \mathbf{u}_{k, m}}{\sum_{m=1}^{M_\sigma} \exp \bigl(\frac{-1}{\lambda} \left[\mathbf{\tilde{S}}\left(\Tau_{m}\right) -\tilde{S}_{\min} \right]\bigr)}\!\!\right)$,
    \EndFor 
    \State $\mathbf{u}_{0} \leftarrow$ \textit{SendToActuators}($\mathbf{U}$),
    \For{$k \leftarrow 1$ \textbf{to} $N-1$}
        \State $\mathbf{u}_{k-1} \leftarrow \mathbf{u}_{k}$,
    \EndFor
    \State \ihab{$\mathbf{u}_{N-1} \leftarrow$ \textit{ControlSequenceInitializer}($\mathbf{x}_\text{init}$),}
    \State Check for task completion
\EndWhile
\end{algorithmic}
\end{algorithm}

\subsection{\textit{Real-Time} U-MPPI Control Algorithm}\label{Pseudo-Code of UMPPI Algorithm}
We are now prepared to describe the \textit{real-time} control-loop of our U-MPPI algorithm, as depicted in Algorithm~\ref{alg:UMPPI-Alg.}, employing the default sampling strategy that considers all sigma points (referred to as $\texttt{SM}_1$). 
\ihab{We assume that the system dynamics, control parameters (including UT parameters), cost function for the given task, and control policy for initializing the control sequence for a given initial system state \(\mathbf{x}_{\text{init}}\) are provided.\footnote{\ihab{In this study, the default initialization policy is applied, producing a control sequence of zeros.}} First,} at every \ihab{\sout{time-step} control loop interval} $\Delta t$, the algorithm estimates the current system state $\Bar{\mathbf{x}}_0$ with an external state estimator, generates $ N \times M_\sigma$ random control perturbations $\delta \mathbf{u}$ on the GPU using CUDA's random number generation library, and then produces $M_\sigma$ sets of batches in parallel on the GPU (lines $2:4$). 
 Subsequently, for each batch and starting from the actual (i.e.,  initial) state $\mathbf{x}_{0} \sim \mathcal{N}(\Bar{\mathbf{x}}_0, \mathbf{\Sigma}_0)$, the algorithm samples and propagates $n_\sigma$ sigma-point trajectories by applying the non-linear dynamics in \eqref{eq:dynamics of ALL sigma points} to the sigma points computed via \eqref{eq:ut-sigma-points} (lines $5:9$). These trajectories are then evaluated using \eqref{eq:cost-to-go-umppi}, and the first and second moments of the propagated state are estimated by applying \eqref{eq:ut-mean-cov} (lines $10:15$). 
 Then, the algorithm updates the optimal control sequence $\left\{\mathbf{u}_{k}\right\}_{k=0}^{N-1}$, applies a Savitzky-Galoy filter for smoothing, and applies the first control $\mathbf{u}_{0}$ to the system (lines $16:19$). It then slides down the remaining sequence of length $N - 1$ to be utilized in the next control loop interval (lines $20:22$). \ihab{Lastly, the final control input \(\mathbf{u}_{N-1}\) is assigned the value determined by the predefined initialization control policy (line 23).}
It is noteworthy that when executing the algorithm in sampling mode 0 ($\texttt{SM}_0$), where only the mean $\mathcal{X}_k^{(0)}$ is used for trajectory sampling, it is essential to set $M_\sigma$ to $M$ instead of $int(\frac{M}{n_\sigma})$. Additionally, it is necessary to evaluate solely the \textit{cost-to-go} of the nominal trajectory $\tilde{S}\left(\tau^{(0)}_{m}\right)$.  
\ihab{\subsection{Constraint Handling and Scalability in U-MPPI}}
\ihab{Herein, we explore two critical aspects of the U-MPPI control strategy: the effective handling of constraints within MPPI-based control frameworks, including our proposed control strategy, and the scalability and adaptability of the U-MPPI algorithm to higher-dimensional robotic systems, highlighting the advantages and disadvantages of the algorithmic structure given in Algorithm~\ref{alg:UMPPI-Alg.}.}
\subsubsection{\ihab{Constraint Handling for MPPI Variants}}
\ihab{One of the primary advantages of MPPI variants, including our proposed control strategy, is their ability to include costs in the form of \textit{large-weighted} indicator terms, as they do not require explicit gradient computations. These terms provide \textit{impulse-like} penalties when constraints are violated, effectively handling task-related state constraints and collision avoidance constraints as soft constraints. By incorporating appropriate terms, such as indicator functions, into the running cost function 
$\tilde{q}$ in \eqref{cost-to-go-mppi} and \eqref{eq:cost-to-go-umppi}, trajectories that violate these constraints can be effectively penalized.
Control constraints, such as actuator limits, which are considered hard constraints that must not be violated, can similarly be integrated into the running cost. However, a common issue with this approach is that control constraints are treated as soft constraints, making it challenging to ensure that the control input consistently remains within its allowed bounds, even after rejecting trajectories that exceed the control input limits. 
Instead, in this study, control constraints are integrated directly into the underlying non-linear dynamics described in~\eqref{eq: underlying non-linear dynamics} and~\eqref{eq:dynamics of ALL sigma points}. Consequently, the system dynamics are represented as \(\mathbf{x}_{k+1} = f\big(\mathbf{x}_{k}, g(\mathbf{w}_{k})\big)\) and \(\mathbf{X}_{k+1} = \mathbf{f}\left(\mathbf{X}_{k}, g(\mathbf{w}_{k})\right)\), where \(g(\mathbf{w}_{k})\) is an element-wise clamping function that confines the control input, \(\mathbf{w}_{k} = \mathbf{u}_{k} + \delta \mathbf{u}_{k}\), within specified bounds for all samples drawn from the dynamic system. Specifically, \(g(\mathbf{w})\) is defined as \(g(\mathbf{w}) = \max \big(\mathbf{w}_{\min}, \min(\mathbf{w}, \mathbf{w}_{\max})\big)\), with \(\mathbf{w}_{\min}\) and \(\mathbf{w}_{\max}\) denoting the lower and upper control input limits, respectively.
This method effectively converts the optimization problem with control constraints into an unconstrained problem, ensuring that constraints are not violated and the convergence of the MPPI-based algorithm is not compromised, as \(g(\mathbf{w})\) only influences the dynamics.
Further details on system constraints, including other practical concerns, can be found in \cite{williams2018information, mohamed2021mppi}.}
\subsubsection{\ihab{Scalability and Adaptability of U-MPPI}}\label{Scalability and Adaptability of U-MPPI}
\ihab{Our investigation centers on the question: How effectively can the U-MPPI algorithm scale to accommodate robotic systems with higher-dimensional state spaces, such as those in unmanned aerial vehicles (UAVs) and mobile manipulators? We assert that the proposed U-MPPI algorithm effectively scales and adapts to systems with higher degrees of freedom (DoFs) beyond the 3-DoF mobile robot used in our current experiments for the following reasons.
First, the core principles of U-MPPI, including the use of the UT for trajectory sampling and the incorporation of risk-sensitive cost functions, are inherently scalable and adaptable to systems with higher DoFs. The UT, a non-linear transformation method, is designed to efficiently handle higher-dimensional state spaces, making it a superior choice over methods like Gaussian Processes (GP), which require substantially more computational resources. For systems such as UAVs and mobile manipulators, the state space includes additional dimensions such as altitude, pitch, roll, joint angles, and velocities. Our UT-based sampling strategy can extend to these dimensions without significant modifications to the algorithmic structure. }

\ihab{Second, in prior work \cite{mohamed2020model}, we demonstrated the applicability of the MPPI algorithm in achieving real-time autonomous navigation of a UAV with a higher-dimensional state space comprising 12 dimensions, and we anticipate that U-MPPI will achieve a comparable level of performance. However, the increased configuration space might impose a greater computational burden due to the need for additional sigma points. This increased burden highlights the importance of several factors to ensure real-time performance, such as the computational capabilities of the GPU, the number of threads utilized, and the efficiency of the implementation. Our GPU-based implementation is designed to exploit parallel processing, allowing for efficient handling of increased state dimensionality. 
Additionally, the number of sampled trajectories, the chosen prediction time horizon, the optimization of the computational pipeline, and the use of CUDA built-in functions and libraries all play crucial roles in achieving real-time performance.
Memory management, data transfer rates between CPU and GPU, and the optimization of kernel functions are also critical factors impacting the algorithm's efficiency.}

\ihab{Additionally, we have investigated the comparative advantages and disadvantages of parallelizing sigma-point trajectories using a reduced number of threads relative to the traditional MPPI approach, where each thread computes the dynamics and costs for an entire trajectory \cite{williams2017model}. As delineated in Algorithm~\ref{alg:UMPPI-Alg.}, our approach involves each thread processing a batch of sigma-point trajectories, thereby reducing computational overhead as fewer threads result in less context-switching and more efficient utilization of GPU resources. This efficiency accounts for U-MPPI's execution times being nearly equivalent to those of the MPPI algorithm, as evidenced in the results section. However, a notable disadvantage of our algorithmic structure is the increased latency in generating each trajectory, attributed to the elevated workload per thread. This issue is worsened in higher-dimensional state spaces or mathematical functions with exponential computational complexity, which can significantly amplify the computational load per thread. To mitigate this issue, a potential solution is to adopt a per-sigma-point trajectory threading model, akin to the MPPI algorithm, which would distribute the workload more evenly across the available threads, potentially reducing latency and enhancing overall performance.
In conclusion, while our current work provides a foundational evaluation using a 3-DoF mobile robot, we are confident in the U-MPPI algorithm's potential to scale to higher-dimensional robotic systems.
}

\section{Simulation-Based Evaluation}\label{Simulation Details and Results}
In this section, we evaluate the effectiveness of our proposed control strategy by comparing it to the standard MPPI control framework. 
Herein, we focus on two goal-oriented autonomous ground vehicle (AGV) navigation tasks in 2D cluttered environments.
The first task, presented in Section~\ref{Task1:Cluttered Environments}, involves maneuvering the AGV within a given map.
This experiment allows us to understand the algorithmic advantages of the proposed U-MPPI control scheme without adding real-world complexity.
Section~\ref{Task2: Unknown Environments} introduces a more complex and realistic scenario where the map is unknown a priori. 
It examines the adaptability and robustness of our proposed control strategy, providing a thorough evaluation of its potential in real-world applications.

\subsection{Aggressive Navigation in Known Cluttered Environments}\label{Task1:Cluttered Environments}
\subsubsection{Simulation Setup}\label{Simulation Setup:Cluttered Environments}
\ihab{In this study, we employ the kinematics model of a differential drive robot, as specified in \eqref{eq:differential-drive-robot}, for sampling trajectories in the conventional MPPI and propagating sigma points in the proposed U-MPPI method, as outlined in \eqref{eq:dynamics of sigma points}.
The state of the model encompasses the robot's position and orientation in the world frame $\mathcal{F}_{\!o}$, given by $\mathbf{x} = [{x}, {y}, \theta]^{\top}  \in \mathbb{R}^{3}$.
The control input consists of the robot's linear and angular velocities, denoted by $\mathbf{u} = [v,\omega]^{\top}  \in \mathbb{R}^{2}$.}
\ihab{\begin{equation}\label{eq:differential-drive-robot}
\left[\begin{array}{c}
\dot{x} \\
\dot{y} \\
\dot{\theta}
\end{array}\right]=\left[\begin{array}{cc}
\cos \theta & 0 \\
\sin \theta & 0 \\
0 & 1
\end{array}\right]\left[\begin{array}{l}
v \\
\omega
\end{array}\right].
\end{equation}}

To ensure a fair comparison, both MPPI and U-MPPI simulations were conducted under consistent parameters. A time prediction of \SI{8}{\second} and a control rate of \SI{30}{\hertz} were employed, resulting in \num{240} control time-steps (i.e., $N=240$). 
At each time-step $\Delta t$, a total of \num{2499} rollouts were sampled, accompanied by an exploration noise of $\nu=1200$. To account for control weighting, a control weighting matrix $R$ was utilized, which was formulated as $\lambda \Sigma_\mathbf{u}^{-\frac{1}{2}}$.
Additionally, the inverse temperature parameter was set to $\lambda = 0.572$, while the control noise variance matrix $\Sigma_\mathbf{u} = \operatorname{Diag}\left(\sigma_v^2, \sigma_w^2\right)$ was defined as $\Sigma_\mathbf{u} = \operatorname{Diag}\left(0.023, 0.028\right)$.
To smooth the control sequence, we utilized the Savitzky-Galoy (\textit{SG}) convolutional filter with a quintic polynomial function, i.e., $n_{sg}=5$, and a window length $l_{sg}$ of $61$. 
As described in Section~\ref{Unscented-Based Sampling Strategy}, U-MPPI has additional parameters in the unscented transform to regulate the spread of the sigma points.
These parameters are set to $\alpha= 1, k_\sigma= 0.5, \text{ and}, \beta=2$. Additionally, the initial state covariance matrix $\mathbf{\Sigma}_0$ is set to $0.001\mathbf{I}_3$.
\ihab{It is worth mentioning that \(\lambda\) and \(\Sigma_\mathbf{u}\), which play crucial roles in determining the behavior of sampling-based MPC schemes, are set based on the intensive simulations conducted in \cite{mohamed2022autonomous}. These simulations considered various values of \(\lambda\) and \(\Sigma_\mathbf{u}\) with the objective of selecting the optimal set of hyperparameters that adhere to the control constraints. Other parameters were fine-tuned using a \textit{trial-and-error} method, guided by an understanding of the role and impact of each parameter on the algorithm's performance \cite{williams2018information, mohamed2020model, van2004sigma}.}
The baseline and U-MPPI are implemented using Python and incorporated within the Robot Operating System (ROS) framework.
They are executed in \textit{real-time} on an NVIDIA GeForce GTX 1660 Ti laptop GPU.

In the context of the 2D navigation task, MPPI employs a commonly-used instantaneous state-dependent cost function described in \eqref{eq:state-dep-cost-function-mppi}. This cost function consists of two terms. The first term, denoted as $q_{\text{state}}(\mathbf{x}_k)$, encourages the robot to reach the desired state; its formulation, which employs a quadratic expression, is provided in \eqref{eq:quadratic-state}. 
The second term, $q_{\text{crash}}(\mathbf{x}_k)= w_\text{crash} \mathbb{I}_{\text{crash}}$, serves as an indicator function that imposes a high penalty when the robot collides with obstacles, with $\mathbb{I}_{\text{crash}}$ being a Boolean variable and $w_\text{crash}$ representing the collision weighting coefficient.
While implementing the proposed U-MPPI approach, we replace the quadratic cost $q_{\text{state}}(\mathbf{x}_k)$, as depicted in \eqref{eq:state-dep-cost-function-umppi}, with our newly introduced RS cost denoted as $q_{\mathrm{rs}}\left(\mathcal{X}_{k}^{(i)}, \mathbf{\Sigma}_{k}\right)$ and defined in \eqref{Qrs-modified}. 
Within the scope of this work, we set the values of $Q$ and $w_\text{crash}$ as follows: $Q = \operatorname{Diag}(2.5,2.5,2)$ and $w_\text{crash}= 10^3$.
It is worth noting that in this particular task, assigning positive values to $\gamma$ (i.e., $\gamma>0$) ensures a trade-off between compelling the robot to reach its desired state and minimizing the risk of collisions with obstacles. This trade-off arises from the fact that the penalty coefficients matrix $Q_{\mathrm{rs}}$, utilized for tracking the desired state, decreases as the system uncertainty $\mathbf{\Sigma}_{k}$ increases over the time-horizon $N$. 
Therefore, we have chosen $\gamma = 1$ to maintain this trade-off and strike a balance between task completion and collision avoidance.
On the other hand, we believe that assigning negative values to $\gamma$ in applications such as autonomous racing \cite{williams2018information} and visual servoing \cite{mohamed2021mppi} could enhance the performance of U-MPPI. In such tasks, it is crucial to prioritize forcing the current state to reach its desired state, which can be achieved by assigning a higher penalty coefficients matrix $Q_{\mathrm{rs}}$.
\begin{equation}\label{eq:state-dep-cost-function-mppi}
q(\mathbf{x}_k)= q_{\text{state}}(\mathbf{x}_k) + q_{\text{crash}}(\mathbf{x}_k).
\end{equation}
\begin{equation}\label{eq:state-dep-cost-function-umppi}
q\left(\mathcal{X}_{k}^{(i)}, \mathbf{\Sigma}_{k}\right) = q_{\mathrm{rs}}\left(\mathcal{X}_{k}^{(i)}, \mathbf{\Sigma}_{k}\right) 
+ q_{\text{crash}}\left(\mathcal{X}_{k}^{(i)}\right).
\end{equation}
\subsubsection{Simulation Scenario}
In order to assess the effectiveness of the proposed control framework within cluttered environments, three distinct scenarios with different difficulty levels were examined. 
In each scenario, we randomly generate one unique forest type consisting of \num{25} individual forests, resulting in a total of $\mathcal{N}_{T} = 25$ tasks. Each forest represents a cluttered environment with dimensions of $\SI{50}{\meter} \times \SI{50}{\meter}$.
In the first scenario (referred to as \textit{Scenario \#1}), the average distance between obstacles was \SI{1.5}{\meter}, indicated as $d^{\text{obs}}_{\min}=\SI{1.5}{\metre}$; while in the second and third scenarios (i.e., \textit{Scenario \#2} and \textit{Scenario \#3}), they were placed at average distances of \SI{2}{\meter} and \SI{3}{\meter}, respectively.
Additionally, we set the maximum desired velocity $v_{\max}$ of the robot based on the degree of clutter in each scenario.
Specifically, $v_{\max}$ is set to \SI{2}{\meter/\second}, \SI{3}{\meter/\second}, and \SI{4}{\meter/\second} in \textit{Scenario \#1, \#2}, and \textit{\#3}, respectively.
\subsubsection{Performance Metrics}\label{Performance Metrics}
To achieve a fair comparison between the two control strategies, we use the following criteria: 
(i) firstly, in all simulation instances, the robot is required to reach the designated desired pose, denoted as $\mathbf{x}_{f}= [50,50,0]^{\top}$, from a predetermined initial pose $\mathbf{x}_{0}= [0,0,0]^{\top}$, measured in $([\si{\metre}], [\si{\metre}], [\si{\deg}])$; 
(ii) secondly, a comprehensive set of metrics is defined to evaluate the overall performance \cite{mohamed2022autonomous}, including 
the {\em task completion percentage} $\mathcal{T}_{\text{c}}$, \ihab{which is computed as the average distance the robot travels towards the desired state to the total planned distance\footnote{\ihab{In this research, the Euclidean distance between $\mathbf{x}_{0}$ and $\mathbf{x}_{f}$ is used to calculate the planned distance.}};} the {\em success rate} $\mathcal{S}_{R}$, \ihab{which is quantified as the ratio of successfully completed tasks to the total tasks undertaken;} the {\em average number of collisions} $\mathcal{N}_{\text{c}}$ \ihab{encountered by the robot during the tasks}; the {\em average number of local minima occurrences} $\mathcal{R}_{\text{lm}}$, \ihab{indicating how often the robot gets trapped in local minima}; the {\em average distance} traversed by the robot $d_{\text{av}}$ to reach the desired state $\mathbf{x}_{f}$ from its initial state $\mathbf{x}_{0}$; 
the {\em average linear velocity} $v_{\text{av}}$ of the robot during the execution of its task in the cluttered environment; 
and the {\em average execution time per iteration} $t_\text{exec.}$ of the control algorithm. 

\ihab{
These metrics are essential for evaluating the efficacy of a control strategy in autonomous vehicle navigation, highlighting its proficiency in generating safe, efficient, and resilient trajectories under varying conditions.
In detail, the task completion percentage $\mathcal{T}_{\text{c}}$ and success rate $\mathcal{S}_{R}$ measure the effectiveness of the approach, with higher values indicating the control strategy's ability to successfully complete tasks and achieve goals under various conditions. Metrics such as the number of collisions $\mathcal{N}_{\text{c}}$ and local minima occurrences $\mathcal{R}_{\text{lm}}$ provide insights into the safety and robustness of the control strategy, with fewer crashes and reduced occurrences of local minima indicating more reliable navigation.
A shorter average distance $d_{\text{av}}$ indicates more efficient trajectory planning, showcasing optimized path generation. Higher average robot speed $v_{\text{av}}$ reflects the ability to maintain a faster pace without compromising safety, crucial for timely task completion in real-world applications. Lower average execution time $t_{\text{exec.}}$ highlights efficiency and responsiveness, ensuring real-time performance in practical scenarios.
}
In this work, successful task completion is defined as the robot reaching the desired pose without colliding with obstacles within a predefined finite time, represented by $\mathcal{T}_{\text{c}} = 100\%$, $\mathcal{N}_{\text{c}} = 0$, and $\mathcal{R}_{\text{lm}} = 0$. Furthermore, in all scenarios, if the robot fails to reach the desired pose within \SI{70}{\second} while successfully avoiding collisions, we classify the simulation episode as reaching a local minimum, indicated by $\mathcal{R}_{\text{lm}} = 1$.
\begin{table}[t]
\small\addtolength{\tabcolsep}{-5pt} 
\setlength\extrarowheight{1pt}
\caption{
Performance comparisons of the two control schemes, with gray cells indicating better performance \ihab{and values for $d_{\text{av}}$, $v_{\text{av}}$, and $t_{\text{exec.}}$ representing the mean values for only the successful tasks.}} 
\centering
\begin{tabular}{|c|| c| c| c| c| c| c| c|}
\hline
Scheme  & $\mathcal{R}_{\text{lm}}$$(\mathcal{N}_{c})$  & $\mathcal{S}_{R}$ [\%] & $\mathcal{T}_{c}$ [\%] & $d_{\text{av}}$ [\si{\metre}]& $v_{\text{av}}$ [\si{\metre/\second}] & $t_{\text{exec.}}$ [\si{\milli\second}]\\
 \hline  
 \hline
 \multicolumn{7}{|c|}{\textit{\textbf{Scenario \#1:}} $v_\text{max} = \SI{2}{\metre/\second}$ \& $d^{\text{obs}}_{\min}=\SI{1.5}{\metre}, \gamma=1, w_\text{crash}= 10^3$} \\
 \hline
 MPPI & 9 (2) & 78 & 92.86 & \cellcolor{gray!20} 75.18 & $1.84 \pm 0.18$ & 9.68  \\
U-MPPI & \cellcolor{gray!20}2 (0) & \cellcolor{gray!20} 96 & \cellcolor{gray!20} 98.78 & 75.34 & $\cellcolor{gray!20}1.85\pm 0.17$ & \cellcolor{gray!20} 9.12 \\ 
 \hline
 \hline
 \multicolumn{7}{|c|}{\textit{\textbf{Scenario \#2:}} $v_\text{max} = \SI{3}{\metre/\second}$ \& $d^{\text{obs}}_{\min}=\SI{2}{\metre}, \gamma=1, w_\text{crash}= 10^3$} \\
 \hline 
 MPPI & 2 (1) & 94 & 98 & \cellcolor{gray!20} 75.31 & $2.49 \pm 0.73$ & 10.03 \\
U-MPPI & \cellcolor{gray!20}0 (0) & \cellcolor{gray!20} 100 & \cellcolor{gray!20} 100 & 75.78 & $ \cellcolor{gray!20}2.53  \pm 0.49 $& \cellcolor{gray!20} 8.87 \\ 
 \hline
 \hline
 \multicolumn{7}{|c|}{\textit{\textbf{Scenario \#3:}} $v_\text{max} = \SI{4}{\metre/\second}$ \& $d^{\text{obs}}_{\min}=\SI{3}{\metre}, \gamma=1, w_\text{crash}= 10^3$} \\
 \hline
 MPPI & \cellcolor{gray!20}  0 (0) & \cellcolor{gray!20} 100 & \cellcolor{gray!20}  100 & 72.19& $ 3.51 \pm 0.74$ & 8.55 \\ 
U-MPPI & \cellcolor{gray!20}  0 (0) & \cellcolor{gray!20} 100 & \cellcolor{gray!20} 100 & \cellcolor{gray!20}  71.98 & $\cellcolor{gray!20} 3.54  \pm 0.59$ & \cellcolor{gray!20}  7.7
\\ 
 \hline
 \end{tabular}
\label{table:intensiveSimulation-Table}
 \vspace*{-2pt} 
\end{table}
\subsubsection{Simulation Results}\label{Simulation Results: 1}
Table \ref{table:intensiveSimulation-Table} presents the performance analysis of the proposed U-MPPI and the baseline MPPI control strategies, considering the three predefined scenarios. For each scenario, two trials were conducted over the 25 individual forests, resulting in a total of 50 tasks ($\mathcal{N}_T = 50$).
In \textit{Scenario \#1}, where $d^{\text{obs}}_{\min}=\SI{1.5}{\metre}$, it is noteworthy that U-MPPI outperforms MPPI. Specifically, U-MPPI achieves a notably higher task completion percentage ($\mathcal{T}_{c} = 98.78\%$) compared to MPPI ($\mathcal{T}_{c} = 92.86\%)$, effectively avoids collisions ($\mathcal{N}_{c} = 0$), mitigates local minimum occurrences ($\mathcal{R}_{\text{lm}}=2$), and achieves a significantly higher success rate ($\mathcal{S}_{R} = 96\%$ vs. $\mathcal{S}_{R} = 78\%$ when MPPI is utilized).
 Furthermore, it successfully navigates the cluttered environment with a slightly improved average linear velocity $v_{\text{av}}$, which exhibits a very low standard deviation and approaches the maximum desired speed $v_{\max}$ of $\SI{2}{\metre/\second}$ (likewise observed in the other two scenarios).
Similarly, in \textit{Scenario \#2}, with a minimum obstacle distance of $d^{\text{obs}}_{\min}=\SI{2}{\metre}$, U-MPPI achieves a perfect task completion rate of $100\%$, outperforming MPPI's $98\%$, as it successfully avoids collisions and local minima, surpassing the baseline MPPI that experienced one collision with obstacles ($\mathcal{N}_{c} = 1$) and encountered two instances of local minima ($\mathcal{R}_{\text{lm}}=2$).

In the least cluttered scenario, \textit{Scenario \#3}, both control strategies effectively complete all assigned tasks while successfully avoiding obstacles in the cluttered environment. 
However, U-MPPI stands out by offering a slightly more direct route towards the desired pose, with the robot traveling an average distance $d_{\text{av}}$ of approximately \SI{71.98}{\meter}, compared to \SI{72.19}{\meter} when utilizing MPPI.
On the contrary, in \textit{Scenarios \#1} and \#2, MPPI demonstrates an enhanced performance in terms of the average distance traveled $d_{\text{av}}$ by the robot when compared to our proposed U-MPPI.
\ihab{Nevertheless, it should be emphasized that the average distance is computed solely on successful task completions, and MPPI has a lower number of successful tasks compared to U-MPPI, which affects the comparison of average distances.
This indicates that U-MPPI outperforms MPPI in terms of overall route completion and reliability, as it is more effective in achieving collision-free navigation.}
In the last column of Table~\ref{table:intensiveSimulation-Table}, despite both control methods ensuring \textit{real-time} performance (since $t_\text{exec.} < \SI{33.33}{\milli\second}$), it is worth emphasizing that the average execution time $t_\text{exec.}$ of our proposed U-MPPI control strategy is slightly shorter than that of MPPI. 
This can be attributed to the parallel implementation of the U-MPPI algorithm on GPU, where each thread is responsible for computing the dynamics and costs of the entire batch when sampling from all sigma points ($\texttt{SM}_1$). On the other hand, the parallel implementation of MPPI, as well as U-MPPI with sampling mode 0 ($\texttt{SM}_0$), employs a single thread to compute each sampled trajectory, resulting in a relatively longer execution time, as evidenced by the intensive simulations in Table~\ref{table:intensiveSimulation-Table}, as well as Tasks \#3 and \#4 in Table~\ref{table:parameters-effects}, where only the mean $\mathcal{X}_k^{(0)}$ is used for trajectory sampling (i.e., $\texttt{SM}_0$).

To summarize, the intensive simulations clearly demonstrate that our U-MPPI method consistently outperforms the baseline MPPI control framework in all tested scenarios, particularly in environments with higher levels of clutter. These remarkable results can be credited to two key factors: the effective utilization of an unscented-based sampling strategy, which provides more flexible and efficient trajectories, and the incorporation of a \textit{risk-sensitive} (RS) cost function that explicitly takes into account risk and uncertainty during the trajectory evaluation process; thanks to the incorporation of these crucial components, our approach ensures a significantly enhanced exploration of the state-space of the controlled system, even while leveraging the same injected Gaussian noise $\delta \mathbf u_k$ into the mean control sequence, effectively reducing the likelihood of being trapped in local minima and yielding a safer and more resilient control system that is suitable for aggressive navigation in highly complex cluttered environments.

\begin{figure}[!t]%
    \centering
    \hspace*{-20pt} \subfloat[U-MPPI: Success rate $\mathcal{S}_{R}$]{\vspace*{1pt}{\includegraphics[scale=0.95]{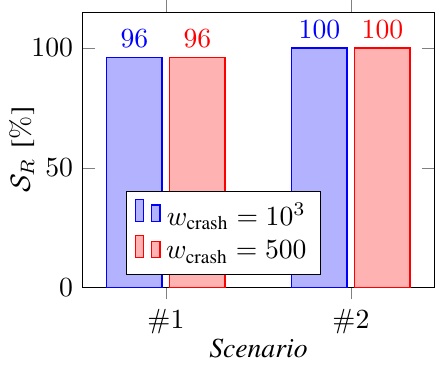}}\label{fig:success-rate-u-mppi-bar}}%
    \subfloat[U-MPPI: Average distance $d_{\text{av}}$]{\vspace*{1pt}{\includegraphics[scale=0.95]{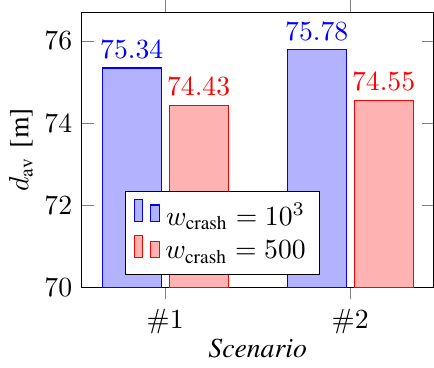}}\label{fig:travelled-distance-u-mppi-bar}}%
    
    \hspace*{-20pt} \subfloat[MPPI: Success rate $\mathcal{S}_{R}$]{\vspace*{1pt}{\includegraphics[scale=0.95]{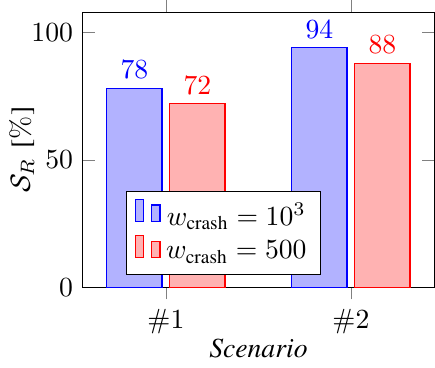}}\label{fig:success-rate-mppi-bar}}%
    \subfloat[U-MPPI vs. MPPI: Success rate $\mathcal{S}_{R}$]{\vspace*{1pt}{\includegraphics[scale=0.95]{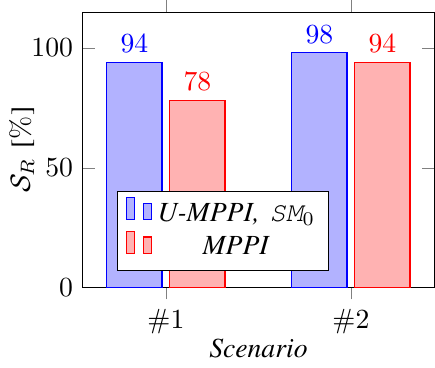}}\label{fig:success-rate-mppi-umppi-bar}}%
    \caption{Impact of decreasing collision weighting coefficient $w_\text{crash}$ on (a) U-MPPI success rate $\mathcal{S}_{R}$, (b) average distance traveled by the robot $d_{\text{av}}$ in U-MPPI, and (c) MPPI success rate $\mathcal{S}_{R}$, as well as (d) the effect of utilizing sampling mode 0 ($\texttt{SM}_0$) on U-MPPI success rate $\mathcal{S}_{R}$ compared to MPPI success rate.
    }
    \label{fig:bar-chart}%
\end{figure}

To achieve a comprehensive understanding of how the behavior of the U-MPPI control strategy is affected by integrating the proposed sampling strategy and RS cost function, we expanded our intensive simulations in Table~\ref{table:parameters-effects} to include varying operating conditions and hyper-parameters, differing from those utilized in Section \ref{Simulation Setup:Cluttered Environments}. 
More precisely, in the first four intensive simulations (namely, Test \#1 to Test \#4), we investigate the potential benefits of integrating the RS cost function into the U-MPPI control strategy through two approaches: (i) reducing the collision weighting coefficient $w_\text{crash}$ (specifically, Tests \#1 and \#2), and (ii) adopting sampling mode 0 ($\texttt{SM}_0$) as an alternative to the default mode $\texttt{SM}_1$ (i.e., Tests \#3 and \#4). Additionally, in the subsequent four tests, we extensively analyze the influence of the UT parameters on the performance of U-MPPI.
For Tests \#1 and \#2, we replicated the U-MPPI simulations presented in Table~\ref{table:intensiveSimulation-Table}, specifically for \textit{Scenarios} \#1 and \#2, by assuming a reduced collision weighting coefficient $w_\text{crash}$ of $500$, representing half of its nominal value.
We can clearly observe that lowering the value of $w_\text{crash}$ has no significant impact on the success rate $\mathcal{S}_{R}$ (as also depicted in Fig.~\ref{fig:success-rate-u-mppi-bar}) and task completion rate $\mathcal{T}_{c}$.  
Nevertheless, it demonstrates improved performance in the robot's average travel distance $d_\text{av}$ for completing the assigned tasks in both scenarios, outperforming both U-MPPI and MPPI as indicated in Table~\ref{table:intensiveSimulation-Table}.
As an example, in \textit{Scenario \#2} shown in Fig.~\ref{fig:travelled-distance-u-mppi-bar}, we can observe that $d_\text{av}$ is approximately \SI{1.23}{\metre} shorter than that of U-MPPI when $w_\text{crash}$ is set to $10^3$.
On the contrary, we empirically observed that reducing $w_\text{crash}$ in the case of MPPI, which utilizes a \textit{risk-neutral} technique for evaluating sampled trajectories (as expressed in \eqref{eq:state-dep-cost-function-mppi}), does not lead to a performance improvement, as depicted in Fig.~\ref{fig:bar-chart}\subref{fig:success-rate-mppi-bar}. 
For instance, in \textit{Scenario \#1}, it can be noted from Fig.~\ref{fig:bar-chart}\subref{fig:success-rate-mppi-bar} that the success rate $\mathcal{S}_{R}$ experiences a decline from 78\% to 72\% with the reduction of $w_\text{crash}$.
\ihab{It is also noteworthy that, although U-MPPI already demonstrated higher average speeds $v_{\text{av}}$ in \textit{Scenarios \#1} and \textit{\#2} in Table~\ref{table:intensiveSimulation-Table}, adjusting $w_\text{crash}$ can further enhance the average speed. In contrast, adopting sampling mode 0 ($\texttt{SM}_0$) and modifying the UT scaling parameters lead to a slight reduction in average speed, as shown in Tests \#3 to \#8.}
\begin{table}[t]
\small\addtolength{\tabcolsep}{-5pt} 
\setlength\extrarowheight{1pt}
\caption{Influence of collision weighting coefficient $w_\text{crash}$, sampling modes (i.e., $\texttt{SM}_0$ and $\texttt{SM}_1$), and UT parameters (namely, $\mathbf{\Sigma}_0, k_\sigma, \alpha$) on U-MPPI performance.} 
\centering
\begin{tabular}{|c|| c| c| c| c| c| c| c|}
\hline
Test No.  &  $\mathcal{R}_{\text{lm}}$$(\mathcal{N}_{c})$ & $\mathcal{S}_{R}$ [\%] & $\mathcal{T}_{c}$ [\%] & $d_{\text{av}}$ [\si{\metre}]& $v_{\text{av}}$ [\si{\metre/\second}] & $t_{\text{exec.}}$ [\si{\milli\second}]\\
 \hline \hline
 \multicolumn{7}{|c|}{\textit{\textbf{Scenarios \#1 and \#2:}} $w_\text{crash}= 500$  instead of $w_\text{crash}= 10^3$} \\
\hline \hline
 Test \#1 & 2 (0) & 96 & 99.4 & 74.43 & $1.86 \pm 0.15$ & 8.69 \\ 
 Test \#2 & 0 (0) & 100 & 100 & 74.55 & $2.56 \pm 0.56 $ & 8.9  \\ 
 \hline \hline
  \multicolumn{7}{|c|}{\textit{\textbf{Scenarios \#1 and \#2:}} $\texttt{SM}_0$ instead of $\texttt{SM}_1$} \\
  \hline 
 \hline
 Test \#3 & 3 (0) & 94 & 98.24 & 75.88 & $1.84 \pm 0.22$ & 12.87 \\ 
 Test \#4 & 1 (0) & 98 & 99.8 & 76.87 & $2.47 \pm 0.67$ & 12.59  \\ 
 \hline
 \hline
 \multicolumn{7}{|c|}{\textit{\textbf{Scenario \#1:}} Impact of UT parameters ($\mathbf{\Sigma}_0, k_\sigma, \alpha$)} \\
\hline 
 \hline 
 Test \#5 & 0 (0) & 100 & 100 & 77.19 & $1.79 \pm 0.23$ & 8.80  \\ 
 Test \#6 & 1 (0) & 98 & 98.7 & 74.65 & $1.80 \pm 0.24$ & 8.88  \\ 
 Test \#7 & 4 (0) & 92 & 96.88 & 75.75 & $1.81 \pm 0.25$ & 9.46  \\ 
  Test \#8 & 9 (0) & 82 & 92.91 & 75.25 & $1.83 \pm 0.22$ & 9.02  \\ 
  \hline
 \end{tabular}
\label{table:parameters-effects}
\end{table}

\begin{figure}[!t]
    \vspace*{-4pt}
    \subfloat[Robot trajectories comparison between MPPI and U-MPPI with random obstacles (black dots)]{{\hspace*{0pt}\includegraphics[scale=1]{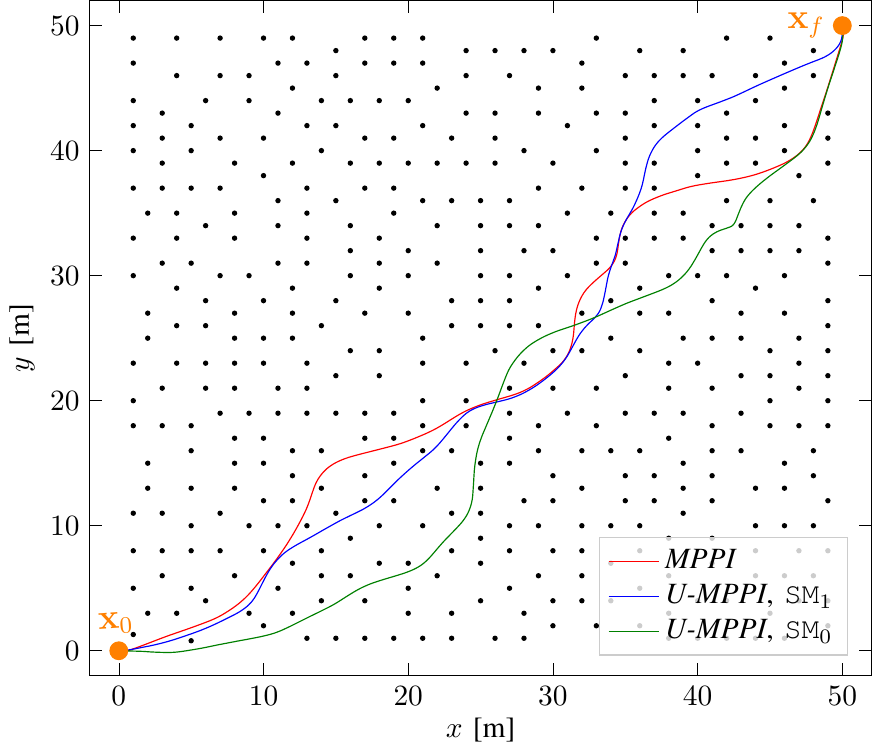}}\label{fig:obs_map}}\\
    \par\medskip 
    \vspace*{-1pt}
    \hspace*{2pt}\subfloat[MPPI: Robot velocity $v$]{\hspace*{2pt}{\includegraphics[scale=1]{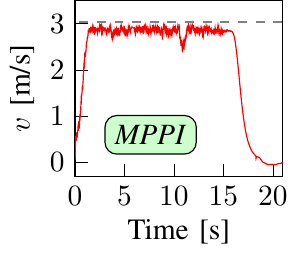}}\label{fig:vel-mppi}}
     \subfloat[U-MPPI, $\texttt{SM}_1$: $v$]{\hspace*{-3pt}{\includegraphics[scale=1]{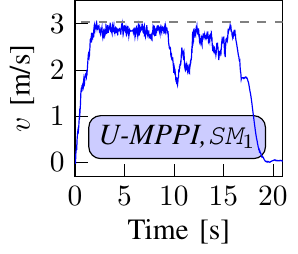}}\label{fig:vel-umppi-sm1}}
     \subfloat[U-MPPI, $\texttt{SM}_0$: $v$]{\hspace*{-3pt}{\includegraphics[scale=1]{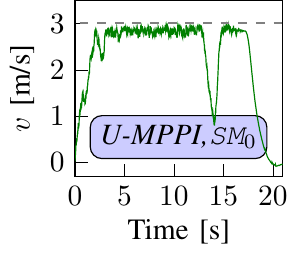}}\label{fig:vel-umppi-sm0}}%
    \caption{Performance analysis of MPPI and U-MPPI in a $\SI{50}{\metre} \times \SI{50}{\metre}$ cluttered environment with \SI{2}{\metre} obstacle spacing (\textit{Scenario \#2}), utilizing U-MPPI-specific sampling modes ($\texttt{SM}_0$ and $\texttt{SM}_1$).}%
    \label{fig:example1}%
\end{figure}

By employing sampling mode 0 ($\texttt{SM}_0$) in Tests \#3 and \#4 as an alternative to the default sampling $\texttt{SM}_1$ in U-MPPI, a slight decrease in performance is observed. However, U-MPPI continues to demonstrate impressive capabilities in successfully accomplishing assigned tasks and navigating around obstacles, outperforming the classical MPPI, particularly in \textit{Scenario \#1} (refer to Fig.~\ref{fig:success-rate-mppi-umppi-bar}), owing to the integration of our efficient RS cost function for trajectory assessment. 
Furthermore, the comprehensive simulations performed in Tests \#3 and \#4 highlight the importance of employing the default sampling strategy $\texttt{SM}_1$, which takes into account all sigma points, in extremely challenging scenarios. This strategy leads to enhancements in both the success rate and the trajectory quality of the robot, as depicted in the illustrative example presented in Fig.~\ref{fig:obs_map}.
\ihab{It is also important to note that the average execution time $t_{\text{exec.}}$ of our proposed U-MPPI algorithm operating in \(\texttt{SM}_0\) is quite longer than in \(\texttt{SM}_1\). This stems from the fact that the algorithm samples \(M\) batches in parallel, rather than \(M_\sigma\) batches, while using only the nominal trajectory $\mathcal{X}_k^{(0)}$ for trajectory sampling and evaluating the cost-to-go $\tilde{S}\left(\tau^{(0)}_{m}\right)$, as explained in Section~\ref{Pseudo-Code of UMPPI Algorithm}. By sampling \(M\) batches, the U-MPPI algorithm ensures that the total number of \(M\) trajectories remains similar to that in MPPI, thereby allowing for a fair comparison.
Additionally, to accurately calculate the risk-sensitive cost function over the time horizon \(N\), it is necessary to propagate all sigma points, even if only the nominal trajectory $\mathcal{X}_k^{(0)}$ is evaluated at each time-step \(k\). This propagation is required to obtain the covariance $\mathbf{\Sigma}_k$, which is then utilized by the risk-sensitive cost function as illustrated in \eqref{Qrs-modified}. 
}

Now, it is time to delve into investigating how the key UT parameters (namely, $\mathbf{\Sigma}_0, k_\sigma, \alpha$) affect the distribution of sampled rollouts and their subsequent influence on the performance of the U-MPPI algorithm; to achieve this, we will explore their effects in highly cluttered environments, with a specific focus on \textit{Scenario \#1}.
In Test \#5, the initial state covariance matrix $\mathbf{\Sigma}_0$ is increased from $0.001\mathbf{I}_3$ to $0.005\mathbf{I}_3$. It is observed that this increase in $\mathbf{\Sigma}_0$ results in a more conservative yet safer trajectory, with a success rate $\mathcal{S}_R$ of 100\% and an average traveled distance $d_\text{av}$ of \SI{77.19}{\metre}, compared to a success rate of 96\% and an average distance of \SI{75.34}{\metre} when $\mathbf{\Sigma}_0$ is set to $0.001\mathbf{I}_3$, as shown in Table~\ref{table:intensiveSimulation-Table}.
During Tests \#6 and \#7, we adjust the UT scaling parameters ($k_{\sigma}$ and $\alpha$), which control the spread of sigma points; specifically, we increase $k_{\sigma}$ from \num{0.5} to \num{3} in Test \#6 and decrease $\alpha$ from \num{1} to \num{0.1} in Test \#7, while keeping all other parameters constant.
It is noteworthy to observe in Test \#6 that assigning a higher value to $k_{\sigma}$, along with $\alpha=1$,  generates more widely spread trajectories that cover a larger state space. This enables the robot to explore the environment more extensively and find better solutions, while improving the quality of the robot trajectory (as $d_\text{av}=\SI{74.65}{\metre} $), thereby reducing the likelihood of getting trapped in local minima.
Conversely, reducing the spread of sigma points by assigning very low values to $\alpha$ increases the likelihood of the robot becoming trapped in local minima, as illustrated in the simulations conducted during Test \#7.
In Test \#8, we replicated the simulations from Test \#7 using the same reduced UT scaling parameters ($k_{\sigma}=0.5$ and $\alpha=0.1$), while excluding the \textit{risk-sensitive} behavior by setting the \textit{risk-sensitive} parameter $\gamma$ to 0, resulting in a constant penalty coefficients matrix $Q_{\mathrm{rs}}$ that is equal to the weighting matrix $Q$.
Such a setup yields a sampling strategy closely resembling that of MPPI, resulting in the worst performance achieved, albeit only marginally better than the performance of MPPI illustrated in Table~\ref{table:intensiveSimulation-Table}.

In Fig. \ref{fig:example1}, we showcase the behavior of MPPI and U-MPPI (utilizing U-MPPI-specific sampling modes) in one of the randomly generated $\SI{50}{\metre} \times \SI{50}{\metre}$ cluttered environments with \SI{2}{\metre} obstacles apart, i.e., \textit{Scenario \#2}. \ihab{This particular environment was selected due to its classification as one of the most challenging environments within \textit{Scenario \#2}, wherein MPPI demonstrated difficulty in achieving efficient navigation by providing a shorter route towards the goal over the two conducted trials.} 
As shown in Fig.~\ref{fig:obs_map}, both control strategies successfully achieve collision-free navigation in the cluttered environment. However, \ihab{owing to the incorporation of the risk-sensitive cost function and the unscented-based sampling strategy}, U-MPPI in the default sampling mode 1 ($\texttt{SM}_1$) demonstrates a significantly shorter route to the desired pose, with a robot trajectory length $d_\text{av}$ of \SI{75.45}{\metre}, compared to \SI{77.39}{\metre} for the classical MPPI and \SI{78.33}{\metre} for U-MPPI in sampling mode 0 ($\texttt{SM}_0$).
In the given cluttered environment, as depicted in Fig.~\ref{fig:vel-mppi}, MPPI demonstrates a smoother velocity profile compared to U-MPPI (see Figs. \ref{fig:vel-umppi-sm1} and \ref{fig:vel-umppi-sm0}), with an average traveling speed $v_{\text{av}}$ of \SI{2.63}{\metre/\second}, as opposed to $v_{\text{av}} = \SI{2.52}{\metre/\second}$ for U-MPPI ($\texttt{SM}_1$) and $v_{\text{av}} = \SI{2.57}{\metre/\second}$ for U-MPPI ($\texttt{SM}_0$).
\ihab{
 The observed fluctuations in the velocity profile under U-MPPI can indeed be attributed to the increased state-space exploration induced by the UT; specifically, the UT propagates both the mean and covariance of the system state, resulting in a more comprehensive exploration of possible trajectories. Consequently, this leads to more aggressive adjustments in control inputs, which manifest as fluctuations in the velocity profile. Furthermore, the risk-sensitive cost function in U-MPPI, which incorporates the system's uncertainty, amplifies this behavior. Notably, in our U-MPPI implementation, the risk sensitivity parameter \(\gamma\) is set to 1, leading to more conservative control actions, as detailed in Section~\ref{Risk-Sensitive Cost}. This setting prioritizes safety and robustness by explicitly accounting for uncertainty, thereby contributing to the observed velocity fluctuations.
 In summary, the fluctuations in the velocity profile under U-MPPI are a direct consequence of the algorithm's design to handle uncertainties and explore a wider range of trajectories while maintaining conservative control actions. This is particularly evident with \(\gamma > 0\) (e.g., \(\gamma = 1\)) and when navigating at high speed in extremely crowded environments such as \textit{Scenario \#2}. These fluctuations indicate the algorithm's capacity to dynamically adjust to changing conditions, ensuring robust and safe navigation.
 }
Furthermore, it is worth noting that none of the control strategies violate the control (namely, velocity) constraint, which is defined as $v\leq v_{\text{max}}=\SI{3}{\metre/\second}$, as observed from the velocity profiles.
\begin{figure}[t]%
    \centering
    \subfloat[Gazebo forest-like environment]{\vspace*{1pt}{\includegraphics[height=1.1in, width=.5\columnwidth]{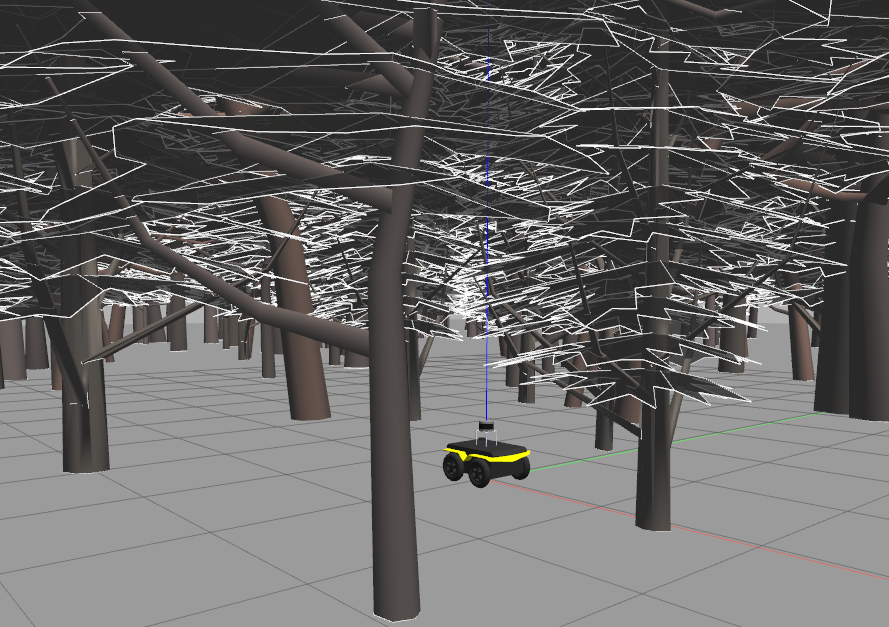}}\label{fig:forst-gazebo}}%
    \;\!\!
    \subfloat[Rviz costmap visualization]{\vspace*{1pt}{\includegraphics[height=1.1in, width=.46\columnwidth]{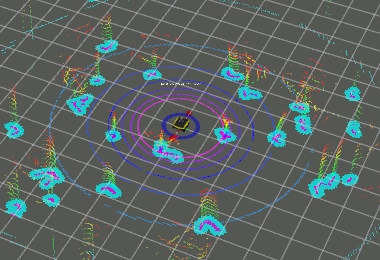}}\label{fig:forst-costmap}}%
    \caption{Snapshot capturing (a) our Jackal robot operating in a forest-like environment, equipped with a Velodyne VLP-16 LiDAR sensor, and (b) the corresponding \textit{2D} local costmap generated by the on-board Velodyne LiDAR sensor, where purple and cyan cells depict obstacles and their inflation. 
    }
    \label{fig:forst-gazebo-rviz}%
\end{figure}

\subsection{Aggressive Navigation in Unknown Environments}\label{Task2: Unknown Environments}
\ihab{\sout{In Section~\ref{Task1:Cluttered Environments}, although extensive simulations of our proposed control strategies are performed using the fully autonomous ClearPath Jackal robot, that the kinematics model outlined in\eqref{eq:differential-drive-robot} to rollout trajectories for the proposed control strategies, to ensure their effectiveness and validate their performance in realistic scenarios, the autonomous navigation tasks in cluttered environments are typically carried out under the assumption that the costmap representing the environment is known beforehand.}
In Section~\ref{Task1:Cluttered Environments}, extensive simulations of our proposed control strategies were conducted using a fully autonomous differential wheeled ClearPath Jackal robot to ensure their effectiveness and validate their performance in realistic scenarios, utilizing the kinematics model outlined in \eqref{eq:differential-drive-robot} for rolling out sampled trajectories for both control strategies. However, it is important to note that the autonomous navigation tasks in cluttered environments are typically carried out under the assumption that the costmap representing the environment is known beforehand.}
Such a setup or configuration could be limited in many real-world scenarios where autonomous robot systems often need to operate in \textit{partially} observed environments due to the constraints of limited sensor range and the impracticality of obtaining a complete map before task deployment~\cite{mohamed2020model}.
To tackle this challenge, the robot needs to construct a \textit{real-time} 2D local costmap, also known as a 2D occupancy grid map, centered at its current state; this grid map is utilized to store information about obstacles in the robot's surrounding area, gathered from the incoming sensory data acquired by its on-board sensor. 
To this end, we utilize the \textit{costmap\_2d} ROS package to generate the occupancy grid based on the sensory data, as depicted in Fig.~\ref{fig:forst-costmap}~\cite{costmap2016}.
Afterward, the local costmap is incorporated into the optimization problem of the sampling-based MPC algorithm to assess sampled trajectories, aiming to achieve collision-free navigation.
We refer to our previous work \cite{mohamed2022autonomous} and the corresponding open-source code.\footnote{\url{https://github.com/IhabMohamed/log-MPPI_ros}} In this work, we successfully integrated the 2D grid map into the MPPI algorithm.

\begin{table}[t]
\small\addtolength{\tabcolsep}{-5.6pt} 
\setlength\extrarowheight{1pt}
\caption{\ihab{Performance statistics of the two control strategies over ten trials in \textit{Forest \#1} (\SI{0.1}{\trees/\square \metre}) and \textit{Forest \#2} (\SI{0.2}{\trees/\square \metre}).}}
\centering
\begin{tabular}{|c||c|c|c|c|c|c|c|}
\hline
Scheme & $\mathcal{R}_{\text{lm}}(\!\mathcal{N}_{c}\!)$ & $\mathcal{S}_{R}$ \!\![\%] & $\mathcal{T}_{c}$ \!\![\%] & $d_{\text{av}}$ [\si{\metre}] & $v_{\text{av}}$ [\si{\metre/\second}] & $t_{\text{exec.}}$ [\si{\milli\second}] \\
\hline  
\hline
\multicolumn{7}{|c|}{\textit{\textbf{Forest \#1:}} $v_\text{max} = \SI{3}{\metre/\second}$ \& $w_\text{max} = \SI{3}{\radian/\second}, \gamma=1, w_\text{crash}= 10^3$} \\
\hline
MPPI & 2 (0) & 80 & 91.01 & \cellcolor{gray!20}185.33$\pm$0.43 & \cellcolor{gray!20} 2.50$\pm$0.68 & \cellcolor{gray!20}8.52$\pm$0.61 \\
U-MPPI & \cellcolor{gray!20}0 (0) & \cellcolor{gray!20}100 & \cellcolor{gray!20}100 & 185.37$\pm$0.22 &2.40$\pm$0.75 & 9.87$\pm$0.76 \\
\hline
\hline
\multicolumn{7}{|c|}{\textit{\textbf{Forest \#2:}} $v_\text{max} = \SI{2}{\metre/\second}$ \& $w_\text{max} = \SI{1}{\radian/\second}, \gamma=1, w_\text{crash}= 10^3$} \\
\hline
MPPI & 4 (1) & 50 & 88.08 & 118.25$\pm$1.7 & 1.56$\pm$0.44 & \cellcolor{gray!20}8.08$\pm$0.91 \\
U-MPPI & \cellcolor{gray!20}1 (0) & \cellcolor{gray!20}90 & \cellcolor{gray!20}97.91 &\cellcolor{gray!20} 116.76$\pm$0.43 & \cellcolor{gray!20}1.67$\pm$0.41 & 10.75$\pm$0.75 \\
\hline
\end{tabular}
\label{table:Comparison-unknowenviroments}
\vspace*{-2pt} 
\end{table}

\subsubsection{Simulation Setup}\label{Simulation Setup:Unknown Environments}
We employ the same simulation setup for both control strategies previously presented in Section~\ref{Simulation Setup:Cluttered Environments}, with the exception that: (i) the collision indicator function $q_{\text{crash}}$ is herein calculated based on the robot-centered 2D grid map, and (ii) the time prediction is reduced to \SI{6}{\second}, resulting in $N=180$, to be compatible with the size of the grid map.
In this study, a 16-beam Velodyne LiDAR sensor mounted on the Clearpath Jackal AGV is used to construct the grid map (local costmap), with the costmap having dimensions of $\SI{240}{\cell} \times \SI{240}{\cell}$ and a resolution of \SI{0.05}{\metre/\cell}.
\subsubsection{Simulation Scenarios}
For the performance evaluation, we utilize two types of forest-like cluttered environments within the Gazebo simulator, each measuring $50\mathrm{m} \times 50\mathrm{m}$. The first type, referred to as \textit{Forest \#1}, consists of trees of different sizes with a density of \SI{0.1}{\trees/\square \metre}, whereas the second type, known as \textit{Forest \#2}, contains tree-shaped obstacles with a density of \SI{0.2}{\trees/\square \metre}.
In \textit{Forest \#1}, the maximum desired velocity $v_\text{max}$ is \SI{3}{\metre/\second}, and the maximum desired angular velocity $w_\text{max}$ is \SI{3}{\radian/\second}. While in \textit{Forest \#2}, they are set to \SI{2}{\metre/\second} and \SI{1}{\radian/\second}, respectively.

\subsubsection{Performance Metrics}
To assess the efficiency of the U-MPPI control scheme in unknown environments, we compare it with the baseline MPPI by evaluating the predefined set of performance metrics outlined in Section~\ref{Performance Metrics}, namely: $\mathcal{R}_{\text{lm}}$, $\mathcal{N}_{c}$, $\mathcal{S}_{R}$,
$\mathcal{T}_{c}$, $d_{\text{av}}$, $v_{\text{av}}$, and 
$t_{\text{exec.}}$.
In the first forest-like environment (\textit{Forest \#1}), the robot is directed to autonomously navigate from an initial pose $G_0= [0,0,0]^\top $ to a sequence of desired poses expressed in ([\si{\metre}], [\si{\metre}], [\si{\deg}]): $G_1= [20,20,45]^{\top}, G_2= [-18,2,0]^{\top}, G_3= [20,-21, 90]^{\top}$, $G_4= [20,20,0]^{\top}$, and ultimately reaching a stop at $G_5= [0,0,100]^{\top}$. Meanwhile, in \textit{Forest \#2}, for the sake of simplicity, the robot navigates solely from $G_0$ to $G_3$, where it comes to a stop.

\subsubsection{Simulation Results}
In Table~\ref{table:Comparison-unknowenviroments}, we present a comparison of performance statistics for the proposed control strategies in achieving goal-oriented autonomous navigation in both \textit{Forest \#1} and \textit{Forest \#2}, where the statistics are averaged over $10$ trials for each environment.
The obtained results validate the anticipated superiority of U-MPPI over MPPI in all scenarios, owing to its efficient unscented-based sampling distribution policy and \textit{risk-sensitive} based trajectory evaluation technique. This superiority is evident in several aspects, including: (i) achieving a higher task completion rate $\mathcal{T}_{c}$, such as $\mathcal{T}_{c}=100\%$ with U-MPPI versus 91.01\% when using MPPI in \textit{Forest \#1}, (ii) reducing the probability of encountering local minima, as demonstrated by the comparison in \textit{Forest \#2}, where U-MPPI leads to $\mathcal{R}_{\text{lm}} = 1$ compared to $\mathcal{R}_{\text{lm}}=4$ with MPPI, and (iii) improving the quality of the generated robot trajectory, as evidenced by a significantly shorter average distance traveled by the robot $d_{\text{av}}$, particularly noticeable in \textit{Forest \#2}.
Furthermore, it is worth emphasizing that both control methods guarantee \textit{real-time} performance, highlighting the superiority of the sampling-based MPC algorithm, particularly our proposed U-MPPI, in incorporating not only the local costmap but also the unscented transform into the optimization problem without introducing additional complexity.

\begin{figure}[!t]
    \centering
    \hspace{-1pt}\subfloat[MPPI: @ $\!G_2$]{{\includegraphics[width=52.5pt,height=45.29pt]{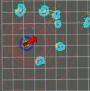}}\label{fig:mppi@G2}}%
    \;
    \subfloat[After \SI{1}{\second}]{{\includegraphics[width=52.5pt,height=45.29pt]{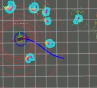}}}%
     \;
    \subfloat[After \SI{2}{\second}]{{\includegraphics[width=52.5pt,height=45.29pt]{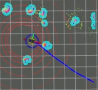}}}%
     \;
    \hspace*{-6pt} \subfloat[After \SI{3}{\second}]{\vspace*{-3pt}{\includegraphics[width=59pt,height=52pt]{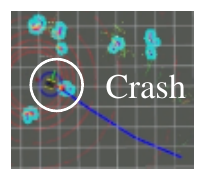}}\label{fig:mppi@G2-last}}\\
    \vspace*{1pt}
    \hspace*{-3pt}\subfloat[Ours: @ $\!G_2$]{{\includegraphics[width=52.5pt,height=45.29pt]{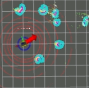}}\label{fig:umppi@G2}}%
    \;
    \subfloat[After \SI{1}{\second}]{{\includegraphics[width=52.5pt,height=45.29pt]{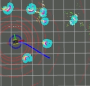}}}%
     \;
    \subfloat[After \SI{2}{\second}]{{\includegraphics[width=52.5pt,height=45.29pt]{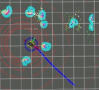}}}%
     \;
    \subfloat[After \SI{3}{\second}]{{\includegraphics[width=52.5pt,height=45.29pt]{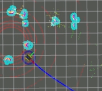}}\label{fig:umppi@G2-last}}
    \vspace*{3pt}
    \caption{\ihab{Planned trajectories by MPPI (first row) and U-MPPI (second row) using the 2D local costmap, starting from the updated pose $G_2= [-18,2,40]^{\top}$ and navigating towards $G_3= [20,-21, 90]^{\top}$.}}
    \label{fig:updated_G2}
\end{figure}

To showcase the superior performance of U-MPPI compared to conventional MPPI in enhancing control safety during aggressive navigation, we introduce a specific adjustment to the $G_2$ configuration within \textit{Forest \#1}. This adjustment entails shifting the desired heading angle from 0 to 40 degrees (see Fig.~\ref{fig:mppi@G2}), resulting in $G_2 = [-18, 2, 40]^{\top}$ instead of $G_2 = [-18, 2, 0]^{\top}$, which in turn requires the robot to perform additional rotational maneuvers.
Throughout five simulation runs, our proposed approach effectively completes all trials while penalizing trajectories with $w_{\text{max}}$ values exceeding \SI{3}{\radian/\second}. Figures~\ref{fig:umppi@G2} to ~\ref{fig:umppi@G2-last} depict snapshots of collision-free and predicted optimal trajectories generated by U-MPPI at three successive time points as the robot advances towards its target pose, $G_3$.
On the other hand, when MPPI is employed, the robot experiences collisions with the same tree on three separate occasions out of the five trials, as depicted in Fig.~\ref{fig:mppi@G2-last}. It is noteworthy that a lower value of $w_{\text{max}}$ is associated with improved MPPI performance. For instance, setting $w_{\text{max}}$ to \SI{1}{\radian/\second} results in the successful completion of all trials without any collisions. 
These outcomes emphasize the enhanced safety delivered by our U-MPPI method without imposing excessive constraints on control inputs.

\vspace*{-8pt}

\section{Real-World Demonstration}\label{Real-World Demonstration}
In this section, we conduct experiments to demonstrate the effectiveness and practicality of the proposed control strategies in achieving \textit{real-time}, collision-free 2D navigation within an unknown and cluttered indoor environment. 
\subsection{\ihab{Navigation in Unknown Corridor Environment}}
\subsubsection{Experimental Setup and Validation Environment}\label{Experimental Setup:real-world Environment}
In our experimental configuration, we utilize an autonomous Clearpath Jackal robot equipped with a 16-beam Velodyne LiDAR sensor, which serves the purposes of generating the local costmap and estimating the robot's pose through the Lidar Odometry and Mapping (LOAM) algorithm \cite{zhang2014loam}.
\begin{wrapfigure}{r}{0.23\textwidth}
\begin{center}
\includegraphics[scale=0.15]{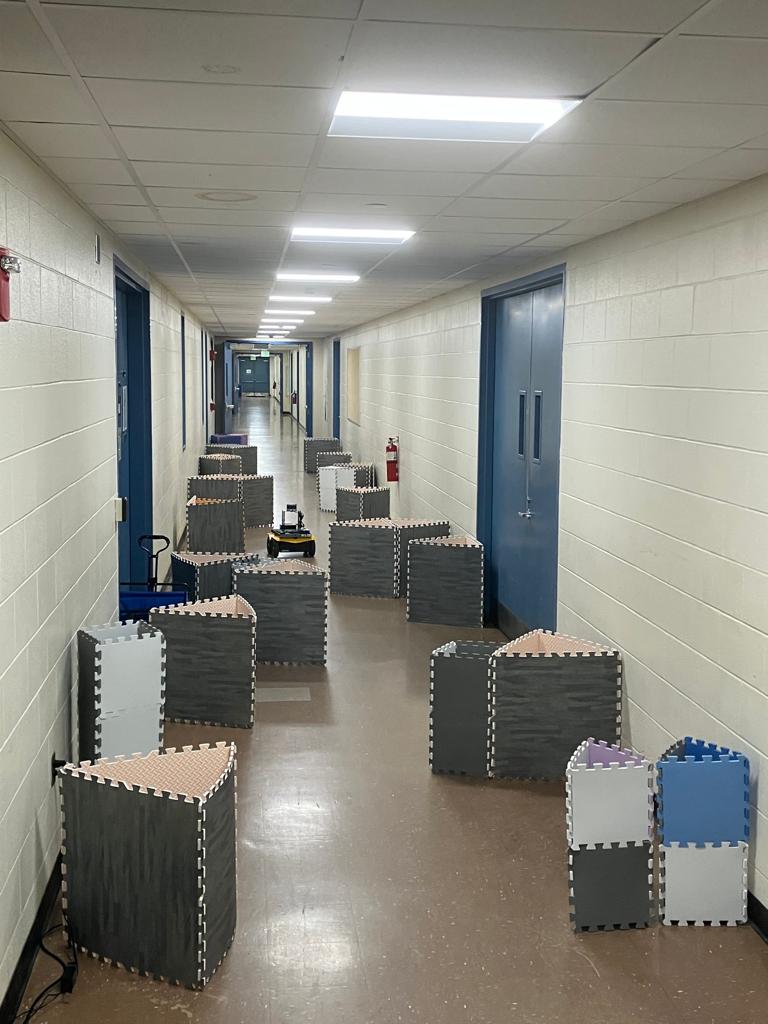}
\caption{Snapshot of our Jackal robot equipped with a Velodyne
VLP-16 LiDAR and located in an indoor unknown cluttered corridor environment.}
\label{fig:corridor-indoor-environment}
\end{center}
\end{wrapfigure}
Additionally, we employ the simulation setup previously outlined in Sections \ref{Simulation Setup:Cluttered Environments} and \ref{Simulation Setup:Unknown Environments}, with two specific modifications: (i) we adjust the maximum speed $v_\text{max}$ to \SI{1.3}{\metre/\second} to mitigate the robot localization error associated with using the Velodyne as a source of localization, and (ii) we extend the time prediction to $\SI{8}{\second}$, an increase from the $\SI{6}{\second}$ duration specified in Section \ref{Simulation Setup:Unknown Environments}. 
We conduct experimental validation in an indoor corridor cluttered environment, measuring $\SI{26}{\metre} \times \SI{2.4}{\metre}$ and containing randomly placed, varying-sized box-like obstacles, as depicted in Fig.~\ref{fig:corridor-indoor-environment}. In this setting, the robot's assigned control mission is to navigate from the following desired poses (in [\si{\metre}], [\si{\metre}], [\si{\deg}]): $G_1= [0,0,0]^T$, $G_2= [26,0.3, 170]^T$, and then return to $G_1$.

\subsubsection{Performance Metrics}
We adhere to the predefined set of performance metrics detailed in Section~\ref{Performance Metrics}. 
Furthermore, given the significant uncertainties and noise inherent in real-world conditions (e.g., caused by LOAM) and the presence of unknown environmental factors, the robot's motion tends to be less smooth than in simulations. To facilitate a more thorough performance evaluation between the two proposed control methods, we opt to calculate the cumulative linear and angular jerks ($\mathrm{J}_{\mathrm{acc}}, \zeta_{\text {acc}}$). 
Jerk, which represents the time derivative of acceleration, is linked to sudden changes in the forces exerted by the vehicle's actuators. Consequently, we can quantify the smoothness of the robot's velocity and steering by examining jerk as a metric \cite{ali2023gp}:
\begin{equation}
\mathrm{J}_{\text{acc}}=\frac{1}{\mathrm{~T}_{\text{tot}}} \int_0^{\mathrm{T}_{\text{tot}}}[\ddot{v}(t)]^2 d t, 
\end{equation}
\begin{equation}
\zeta_{\text{acc }}=\frac{1}{\mathrm{~T}_{\text{tot}}} \int_0^{\mathrm{T}_{\text{tot}}}[\ddot{w}(t)]^2 d t, 
\end{equation} 
where $\mathrm{T}_{\text{tot}}$ represents the complete duration required for the robot to execute its control mission. 
Additionally, we introduce the variable $\mathcal{E}_{\text{loc.}} = [\mathcal{E}_x, \mathcal{E}_y, \mathcal{E}_\theta]^\top$, which signifies the average drift or localization error in the robot's pose relative to the world frame $\mathcal{F}_{\!o}$, as displayed in Fig.~\ref{fig:worst-localization-error}.

\vspace*{5pt}
\begin{table}[t]
\centering
\caption{Performance statistics of the two control strategies over
six trials in an indoor corridor environment.} 
\begin{tabular}{|l||c|c|}
\hline
Indicator  &            MPPI &              U-MPPI\\
\hline
\hline
$\mathcal{R}_{\text{lm}}$ $(\mathcal{N}_{c})$ & 0 (1) & \cellcolor{gray!20} 0 (0) \\
$d_{\text{av}}$ [\si{\metre}] & $ 56.58 \pm 0.33$ & $\cellcolor{gray!20} 56.01 \pm 0.46$ \\
$v_{\text{av}}$ [\si{\metre/\second}] & $1.01 \pm 0.32 $  & $\cellcolor{gray!20} 1.02 \pm 0.28 $ \\
$t_{\text{exec.}}$ [\si{\milli\second}] &  $\cellcolor{gray!20} 9.07 \pm 0.40$ & $10.78 \pm 0.42$ \\
\hline
\hline
$\mathrm{J}_{\text{acc}}$ [\si{\square \metre/\cubic\second}] & $257.75 \pm 62.99$ &  $\cellcolor{gray!20} 228.15  \pm 22.01 $ \\
$\zeta_{\text{acc}}$ [\si{\square \radian/\cubic\second}] &  $306.91  \pm 15.74$   & \cellcolor{gray!20} $263.02  \pm  25.10 $ \\
\hline\hline
$\mathcal{E}_x$ [\si{\metre}] w.r.t. $\mathcal{F}_{\!o}$ &  $0.38 \pm 0.67$ & \cellcolor{gray!20}$0.10 \pm 0.28$ \\
$\mathcal{E}_y$ [\si{\metre}] w.r.t. $\mathcal{F}_{\!o}$ &  $0.30 \pm 0.34$ & \cellcolor{gray!20}$-0.04\pm 0.04$ \\
$\mathcal{E}_\theta$ [\si{\deg}] w.r.t. $\mathcal{F}_{\!o}$ &  $7.20 \pm 11.65$ & \cellcolor{gray!20}$ 0.80\pm 3.71$ \\
\hline
\end{tabular}
\label{table:real-world-results}
\end{table}
\subsubsection{Experimental Results}\label{Experimental results:real-world Environment} 
\begin{figure}[!t]%
    \centering
    \subfloat[Vanilla MPPI]{{
    \includegraphics[scale=1]{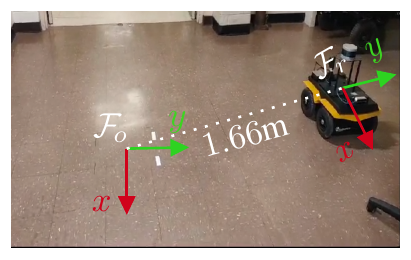}
    }\label{fig:loc-mppi}}%
    \subfloat[Our proposed U-MPPI]{\vspace*{1pt}{
    \includegraphics[scale=1]{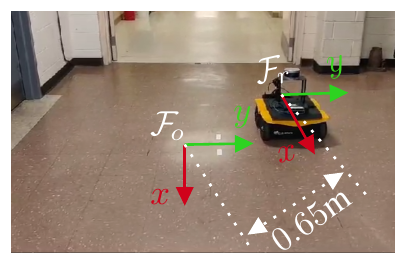}
    }\label{fig:loc-umppi}}%
    \caption{\ihab{Visualization of the cumulative errors in the robot's pose estimation generated by LOAM for the worst trial when employing (a) MPPI and (b) U-MPPI.} 
    \vspace*{-3pt}
    }
    \label{fig:worst-localization-error}%
\end{figure}

The performance statistics comparison between the two control strategies, MPPI and U-MPPI, across six trials in our indoor corridor environment is summarized in Table~\ref{table:real-world-results}.
Throughout all trials, U-MPPI consistently outperforms MPPI in collision avoidance, ensuring collision-free navigation with zero collisions, whereas MPPI averages one collision.
While both control strategies exhibit similar performance in terms of the average distance traveled $d_{\text{av}}$ by the robot, average linear velocities $v_{\text{av}}$, and \textit{real-time} performance guarantee (with $t_\text{exec.} < \SI{33.33}{\milli\second}$), U-MPPI demonstrates a slightly longer average execution time per iteration $t_{\text{exec.}}$ when compared to MPPI, due to its incorporation of both the local costmap and the unscented transform into the U-MPPI optimization problem.
It is noteworthy to observe that U-MPPI showcases smoother motion, as indicated by lower cumulative linear jerk $\mathrm{J}_{\text{acc}}$ at \SI{228.15}{\square \metre/\cubic\second} and angular jerk $\zeta_{\text{acc}}$ at \SI{263.02}{\square \radian/\cubic\second}, whereas MPPI exhibits higher jerk values (\SI{257.75}{\square \metre/\cubic\second}  and \SI{306.91}{\square \radian/\cubic\second}, respectively).
As a result of this enhanced motion smoothness, U-MPPI demonstrates improved robot localization accuracy based on LOAM, with smaller errors in both the $x$-direction ($\mathcal{E}_x = \SI{0.1}{\metre}$) and the $y$-direction ($\mathcal{E}_y = \SI{-0.04}{\metre}$), relative to the reference frame $\mathcal{F}_{\!o}$, compared to MPPI (\SI{0.38}{\metre} and \SI{0.30}{\metre}, respectively). However, MPPI has a slightly higher orientation error, specifically $\mathcal{E}_{\theta} = \SI{7.20}{\deg}$, compared to $\mathcal{E}_{\theta} = \SI{0.80}{\deg}$ when U-MPPI is utilized.
Figure~\ref{fig:worst-localization-error} depicts the accumulated robot pose estimation errors generated by LOAM after completing the control mission for the worst trial among the six conducted trials for both control strategies. Thanks to incorporating uncertainty information throughout the trajectory evaluation process, U-MPPI can avoid aggressive control sequences (motions) that may, in sequence, negatively impact LOAM localization accuracy.
These findings emphasize the benefits of U-MPPI in terms of collision avoidance, motion smoothness, and localization accuracy, all while recognizing the trade-off of a slightly longer execution time.
More details about both simulation and experimental results are included in this video: \url{https://youtu.be/1xsh4BxIrng}.

\subsection{\ihab{Discussion}}
    \ihab{This section offers a thorough analysis of the execution times and performance metrics of our proposed control strategies, emphasizing disparities in U-MPPI performance compared to MPPI and proposing methods to enhance its effectiveness. Additionally, it outlines strategies for dynamic obstacle integration to improve navigation robustness.}
    \subsubsection{\ihab{U-MPPI vs. Vanilla MPPI Execution Times}}
    \ihab{
    Let us elaborate on why integrating the Unscented Transform (UT) and the risk-sensitive cost function does not significantly involve additional computational complexity for the U-MPPI optimization problem, resulting in average execution times $t_{\text{exec.}}$ that are nearly equivalent to, and in some cases better than, those of the MPPI algorithm.
    The primary reasons for this improvement are two-fold. First, the U-MPPI algorithm leverages the UT method, a non-linear transformation method that efficiently handles higher-dimensional state spaces, thereby reducing computational complexity and allowing for parallel processing, all contributing to faster execution times.
    Second, the key strength of our algorithmic structure detailed in Algorithm~\ref{alg:UMPPI-Alg.} is the parallel sampling of \(M_\sigma\) sets of batches on the GPU, with each batch containing \(n_\sigma\) trajectories corresponding to the \(n_\sigma\) sigma points, and each batch employing a single thread. 
    Such an algorithmic structure results in the utilization of fewer threads, specifically $int(\frac{M}{n_\sigma})$, in contrast to the MPPI algorithm, which allocates one thread per trajectory \cite{williams2017model}. Consequently, it reduces computational overhead by minimizing context-switching and efficiently utilizing GPU resources.
    These factors clarify why the average execution times $t_{\text{exec.}}$ of the U-MPPI outperform those of the MPPI algorithm in the intensive simulations presented in Table~\ref{table:intensiveSimulation-Table}.
    This can be traced back to the relatively small size of the obstacle costmap, which did not significantly impact the workload per thread.
    Furthermore, efficient memory management for these maps ensures optimal utilization of the limited GPU memory, further contributing to faster execution times.
    It is also noteworthy in Table~\ref{table:intensiveSimulation-Table} that the execution time decreases with the reduction in the map's size. This is evident in \textit{Scenario \#1}, where the presence of numerous obstacles leads to a larger map size, demanding more memory allocation and management. In contrast, \textit{Scenario \#3} features a smaller map size, leading to less memory usage and consequently faster execution times.
    Nevertheless, it is crucial to recognize that in Tables~\ref{table:Comparison-unknowenviroments} and~\ref{table:real-world-results}, where the local costmap with dimensions of $\SI{240}{\cell} \times \SI{240}{\cell}$ built by the onboard sensor is incorporated into the optimization problem, the average execution times of U-MPPI were slightly longer than those of MPPI. This discrepancy can be attributed to the substantial dimensions of the costmap, which impose a higher workload per thread, thereby increasing latency in trajectory generation and necessitating more memory allocation and management than in the simulations presented in Table~\ref{table:intensiveSimulation-Table}.
    For further details on mitigating the increased workload and ensuring real-time performance, refer to Section~\ref{Scalability and Adaptability of U-MPPI}, which discusses several key factors.
    }
\subsubsection{\ihab{Performance Metrics Evaluation}}
    \ihab{
    Through the intensive simulations conducted in Sections~\ref{Simulation Details and Results} and~\ref{Real-World Demonstration}, we have observed that the metrics of task completion percentage $\mathcal{T}_{\text{c}}$ and success rate $\mathcal{S}_{R}$, number of collisions $\mathcal{N}_{\text{c}}$ and local minima occurrences $\mathcal{R}_{\text{lm}}$, average traveled distance $d_{\text{av}}$ by the robot, and average robot speed $v_{\text{av}}$ are pivotal in evaluating the effectiveness of the proposed control strategy.
    These metrics specifically assess the ability to generate safe, efficient, and robust trajectories for autonomous vehicle navigation in cluttered environments under various conditions.
    The results unequivocally demonstrate that our proposed U-MPPI control strategy outperforms the vanilla MPPI in providing safer and more reliable navigation. This superiority is evidenced by higher success and task completion rates, along with a reduced number of collisions and occurrences of local minima.
    However, one might argue that this improvement does not extend to the average traveled distance $d_{\text{av}}$ and average linear speed $v_{\text{av}}$. 
    While it is true that U-MPPI may not always achieve the shortest path, as observed in \textit{Scenarios \#1} and \textit{\#2} in Table~\ref{table:intensiveSimulation-Table} and \textit{Forest \#1} in Table~\ref{table:Comparison-unknowenviroments}, it should be emphasized that the average distance is based solely on successful tasks, with MPPI having fewer successful tasks than U-MPPI. This affects the comparison of average distances and indicates that U-MPPI outperforms MPPI in terms of overall route completion and reliability, demonstrating greater effectiveness in achieving collision-free navigation. To further enhance U-MPPI's trajectory quality, specifically by reducing the average traveled distance, we have proposed several methods, including:
    \begin{enumerate}[label=(\roman*)]
        \item reducing the collision weighting coefficient $w_\text{crash}$, which has no significant impact on the success rate (see Fig.~\ref{fig:travelled-distance-u-mppi-bar} and Tests \#1 and \#2 in Table~\ref{table:parameters-effects}),
        \item assigning a higher value to $k_{\sigma}$, which also slightly reduced the occurrences of local minima from 2 to 1 in \textit{Scenario \#1} (see Test \#6 in Table~\ref{table:parameters-effects}),
        \item setting  $\gamma <0$, which makes the control strategy more aggressive in attempting to reach the desired state, as explained in Section~\ref{Risk-Sensitive Cost}, thereby improving the quality of the robot trajectory. 
        For instance, in \textit{Scenario \#2}, simulations were conducted with $\gamma=-1$ instead of $\gamma=1$. This adjustment demonstrated improved performance in the robot's average travel distance $d_\text{av}$ (namely, $d_\text{av}=74.8 \si{\metre}$), thereby outperforming both U-MPPI and MPPI, as indicated in Table~\ref{table:intensiveSimulation-Table}. Moreover, this configuration achieved a perfect task completion rate of 100\%. 
    \end{enumerate}   
    }

    \ihab{
    Concerning the second argument regarding the improvement in the average linear velocity, it is noteworthy that the MPPI algorithm slightly outperforms the U-MPPI algorithm in only one scenario: \textit{Forest \#1}, as reported in Table~\ref{table:Comparison-unknowenviroments}. However, it achieves marginally higher average speeds in the remaining tests, as indicated in Tables~\ref{table:intensiveSimulation-Table},~\ref{table:Comparison-unknowenviroments}, and~\ref{table:real-world-results}.
    The limited improvements in average linear velocity are primarily due to the risk-sensitive cost function employed by U-MPPI, which prioritizes safer, more robust trajectories, inherently resulting in more conservative control actions with \(\gamma > 0\). Additionally, the integration of the UT for state propagation allows for comprehensive state-space exploration by considering both the mean and covariance of the system dynamics, leading to the selection of more cautious trajectories that ensure safety but may not be the fastest. This might result in fluctuations in the velocity profile of U-MPPI, especially when navigating at high speed in extremely crowded environments, as discussed at the end of Section~\ref{Simulation Results: 1}.
    }
\subsubsection{\ihab{Dynamic Obstacles Integration}}
    \ihab{While our current study evaluates the U-MPPI control strategy in static environments to establish foundational performance metrics, navigating dynamic environments, particularly those shared with humans or other agents, necessitates enhanced safety measures. Despite achieving zero collisions (\(\mathcal{N}_{\text{c}}\)) in all simulations and real-world validations, the challenge of dynamic obstacle avoidance remains. For instance, our previous work \cite{mohamed2022autonomous} demonstrated successful avoidance of moving agents, such as eight pedestrians, in 2D grid-based navigation scenarios. However, an increase in the number of agents led to a noisier 2D costmap, raising the risk of the vehicle being trapped in local minima and compromising safety. Given our current reliance on a 2D grid for collision avoidance, similar behaviors and safety concerns are anticipated.
    Therefore, to tackle this issue and ensure safety, an alternative solution involves enhancing our control strategy by incorporating formulations similar to equation (17) in \cite{mohamed2021mppi}, which adjusts the cost function to better account for collision risks, or by applying chance constraints \cite{du2011probabilistic} to provide probabilistic guarantees of collision avoidance. These measures aim to ensure safety in highly dynamic scenarios, further improving the reliability of our autonomous navigation system.
    }
\vspace*{-3pt}

\section{Conclusion and Future Work}\label{sec:conclusion}
In this paper, we proposed the U-MPPI control strategy, a novel methodology that enhances the vanilla MPPI algorithm by leveraging the unscented transform for two primary objectives. Firstly, it regulates the propagation of the dynamical system, resulting in a more effective sampling distribution policy that effectively propagates both the mean $\bar{\mathbf{x}}_k$ and covariance $\mathbf{\Sigma}_k$ of the state vector $\mathbf{x}_k$ at each time-step $k$. Secondly, it incorporates a \textit{risk-sensitive} cost function that explicitly accounts for risk or uncertainty throughout the trajectory evaluation process.
Through extensive simulations and real-world demonstrations, we demonstrated the effectiveness of U-MPPI in achieving aggressive collision-free navigation in both known and unknown cluttered environments.
By comparing it to MPPI, our approach accomplished a substantial improvement in state-space exploration while utilizing the same injected Gaussian noise $\delta \mathbf u_k$ in the mean control sequence. As a result, it yielded higher success and task completion rates, effectively minimizing the likelihood of getting trapped in local minima, and enabling the robot to identify feasible trajectories that avoid collisions.
Our future plan involves incorporating chance constraints into the U-MPPI control architecture to effectively address uncertainties in system dynamics and the environment, including moving obstacles, resulting in an enhanced safety and robustness of the control system to handle uncertain conditions, especially in safety-critical applications. \ihab{Moreover, we intend to extend our experimental validation to include UAVs, mobile manipulators, and autonomous ground vehicles with higher-dimensional state spaces. This will provide empirical evidence of the algorithm's performance in these contexts, demonstrating its robustness and efficiency in handling more complex systems.
}

\bibliographystyle{IEEEtran}
\bibliography{references}            

\appendices

\section{Derivative of \textit{Risk-Sensitive} Cost}\label{Appendix Risk Sensitive Cost Function}
Considering the \textit{risk-sensitive} cost defined as:
\begin{equation}\label{eq:rs_given_proof}
q_{\text{rs}}(\mathbf{x}_k) =  -\frac{2}{\gamma} \log \mathbb{E}\left[\exp \left(-\frac{1}{2} \gamma\left\|\mathbf{x}_{k}-\mathbf{x}_{f}\right\|_{Q}^2\right)\right], 
\end{equation}
our objective is to obtain the following formulation:
\begin{equation}\label{eq:Q_rs_proof}
q_{\text{rs}}(\mathbf{x}_k) = \frac{1}{\gamma} \log \operatorname{det}\left(\mathbf{I}+\gamma Q \mathbf{\Sigma}_{k}\right)
+\left\|\mathbf{\Bar{\mathbf{x}}}_{k}-\mathbf{x}_{f}\right\|_{Q_{\mathrm{rs}}}^2,
\end{equation}
where $Q_{\mathrm{rs}} = {\left(Q^{-1}+\gamma \mathbf{\Sigma}_{k}\right)^{-1}}$, $Q$ is a diagonal matrix, and $\mathbf \Sigma_k$ is a symmetric matrix. We assume that the system state $\mathbf{x}_k$ follows a Gaussian distribution $\mathcal{N}(\bar{\mathbf{x}}_k, \mathbf{\Sigma}_k)$, and $\mathbf{x}_f$ represents the reference (desired) state.
To derive the expression for $q_{\text{rs}}(\mathbf{x}_k)$ as presented in \eqref{eq:Q_rs_proof}, we will proceed with a step-by-step derivation as follows.
We will start by expanding the quadratic form in \eqref{eq:rs_given_proof}:
\begin{equation}\label{eq:step1}
\begin{aligned}
q_{\text{rs}}(\mathbf{x}_k) &= -\frac2\gamma \log\mathbb E\left[\exp \left\{-\frac\gamma2\left\|\mathbf x_k - \mathbf x_f\right\|_Q^2\right\}\right]\\ 
&= -\frac2\gamma \log\mathbb E\left[\exp \left\{-\frac\gamma2\left(\mathbf x_k - \mathbf x_f\right)^\top Q\left(\mathbf x_k - \mathbf x_f\right)\right\}\right]. 
\end{aligned}
\end{equation}
Since $\mathbf{x}_k \sim \mathcal{N}(\bar{\mathbf{x}}_k, \mathbf{\Sigma}_k)$, we can assume that $\mathbf{x}_k = \bar{\mathbf{x}}_k + \mathbf{e}_k$, where $\mathbf{e}_k \sim \mathcal{N}(0, \mathbf{\Sigma}_k)$. Therefore, by substituting $\mathbf{x}_k$ in \eqref{eq:step1}, we can express the expectation as follows:
\begin{equation}\label{eq:step2}
\begin{aligned}
q_{\text{rs}} &= -\frac{2}{\gamma}\log \mathbb E\left[\exp\left\{-\frac{\gamma}{2}\overline{\mathbf{X}}_k^\top Q\overline{\mathbf{X}}_k - \gamma \mathbf{e}_k^\top Q\overline{\mathbf{X}}_k -\frac{\gamma}{2}\mathbf{e}_k^\top Q\mathbf{e}_k\right\}\right] \\
&= -\frac{2}{\gamma}\log \mathbb E\left[\exp\left\{-\frac{\gamma}{2}\overline{\mathbf{X}}_k^\top Q\overline{\mathbf{X}}_k\right\}\right. \\
& \qquad \times \left.\exp\left\{-\gamma \mathbf{e}_k^\top Q\overline{\mathbf{X}}_k -\frac{\gamma}{2}\mathbf{e}_k^\top Q\mathbf{e}_k\right\}\right] \\
&=\left\|\overline{\mathbf{X}}_k\right\|_{Q}^{2}-\frac{2}{\gamma} \log \mathbb{E}\left[\exp\left\{-\frac{\gamma}{2}\left\|\mathbf{e}_{k}\right\|_{Q}^{2}-\gamma \mathbf{e}_{k}^\top  Q\overline{\mathbf{X}}_k\right\}\right], 
\end{aligned}
\end{equation}
where $\overline{\mathbf{X}}_k = \bar{\mathbf{x}}_k - \textbf{x}_f$.
Now, to compute the expectation $\mathbb{E}[\cdot]$ in \eqref{eq:step2}, given by
\begin{equation}\label{eq:step3}
-\frac{2}{\gamma} \log \mathbb{E}\left[\exp\left\{ - \frac{\gamma}{2} \mathbf{e}_k^\top  Q \mathbf{e}_k -\gamma \mathbf{e}_k^\top  Q \overline{\mathbf{X}}_k\right\}\right],
\end{equation}
we introduce a new term $\mathbf{y}_k$ to simplify and express the expectation in the quadratic form.
To achieve this, let us define $\mathbf{y}_k = -\gamma \mathbf{K}^{-1}Q\overline{\mathbf{X}}_k$, where $\mathbf{K} = \gamma Q + \Sigma_k^{-1}$. By using this definition, we can accordingly prove that 
\begin{equation}\label{eq:step4}
\begin{aligned}
&\frac{1}{2}\left(\mathbf{e}_k - \mathbf{y}_k\right)^\top \mathbf{K}\left(\mathbf{e}_k - \mathbf{y}_k\right) - \frac{1}{2}\mathbf{y}_k^\top \mathbf{K}\mathbf{y}_k \\
&=
\frac{\gamma}{2}\mathbf{e}_k^\top Q\mathbf{e}_k + \frac{1}{2}\mathbf{e}_k^\top \mathbf \Sigma_k^{-1}\mathbf{e}_k + \gamma \mathbf{e}_k^\top Q\overline{\mathbf{X}}_k.   
\end{aligned}
\end{equation}
Building on the equivalence in \eqref{eq:step4},
we can approximate the expectation in \eqref{eq:step3} by integrating with respect to the distribution of the random variable $\mathbf{e}_k$. Since $\mathbf{e}_k \sim \mathcal{N}(\mathbf{0}, \mathbf{\Sigma}_k)$, we can express it as follows:
\begin{equation}\label{eq:step5}
    \begin{aligned}
        &-\frac{2}{\gamma} \log \mathbb{E}\left[\exp\left\{ - \frac{\gamma}{2} \mathbf{e}_k^\top  Q \mathbf{e}_k -\gamma \mathbf{e}_k^\top  Q \overline{\mathbf{X}}_k\right\}\right] \\
        &= -\frac2\gamma \log \left(\int_{\mathbb R^n} \exp\left\{\frac{-1}{2}\left(\mathbf e_k - \mathbf y_k\right)^\top \mathbf{K}\left(\mathbf e_k - \mathbf y_k\right) \right. \right.\\
        & \qquad + \left. \left. \frac12\mathbf y_k^\top \mathbf{K}\mathbf y_k + \frac{1}{2}\mathbf{e}_k^\top \mathbf \Sigma_k^{-1}\mathbf{e}_k \right\} p(\mathbf e_k) \mathrm d \mathbf e_k\right),
    \end{aligned}
\end{equation}
where $p(\mathbf e_k)  = \frac{1}{\sqrt \mathbf{A}} \exp\left\{-\frac{1}{2}\mathbf{e}_k^\top  \mathbf{\Sigma}_k^{-1} \mathbf{e}_k\right\}$ represents the probability density function of $\mathbf{e}_k$. 
By substituting $p(\mathbf e_k)$ in \eqref{eq:step5} and simplifying it further, we obtain:
\begin{equation}\label{eq:step6}
    \begin{aligned}
        &-\frac{2}{\gamma} \log \mathbb{E}\left[\exp\left\{ - \frac{\gamma}{2} \mathbf{e}_k^\top  Q \mathbf{e}_k -\gamma \mathbf{e}_k^\top  Q \overline{\mathbf{X}}_k\right\}\right] \\
        &= -\frac2\gamma \log \left(\frac{1}{\sqrt \mathbf{A}}\int_{\mathbb R^n} \exp\left\{-\frac{1}{2}\left(\mathbf e_k - \mathbf y_k\right)^\top \mathbf{K}\left(\mathbf e_k - \mathbf y_k\right) \right. \right. \\
        &\qquad +\left. \left. \frac12\mathbf y_k^\top \mathbf{K}\mathbf y_k\right\}\mathrm d \mathbf e_k\right)\\
        &= -\frac2\gamma \log \left(\frac{1}{\sqrt \mathbf{A}} \exp\left(\frac{1}{2}\mathbf{y}_k^\top \mathbf{K}\mathbf{y}_k\right)\right. \\
        & \left. \qquad \times \int_{\mathbb R^n}\exp\left\{\frac{-1}{2}\left(\mathbf e_k - \mathbf y_k\right)^\top \mathbf{K}\left(\mathbf e_k - \mathbf y_k\right)\right\}\mathrm d \mathbf e_k\right), 
    \end{aligned}
\end{equation}
where $\textbf{A} = (2\pi)^n\det \mathbf{\Sigma}_k$. 
We can now evaluate the integral. Since the integrand is a Gaussian distribution with mean $\mathbf y_k$ and covariance matrix $\mathbf{K}^{-1}$, the integral evaluates to $\sqrt{\frac{(2\pi)^n}{\det \mathbf{K}}}$. Substituting this back into the expression given in \eqref{eq:step6}, we have:
\begin{equation}\label{eq:step6'}
\begin{aligned}
&-\frac{2}{\gamma} \log \mathbb{E}\left[\exp\left\{ - \frac{\gamma}{2} \mathbf{e}_k^\top  Q \mathbf{e}_k -\gamma \mathbf{e}_k^\top  Q \overline{\mathbf{X}}_k\right\}\right] \\
&= -\frac2\gamma \log \left(\frac{\exp\left(\frac{1}{2}\mathbf{y}_k^\top \mathbf{K}\mathbf{y}_k\right)}{\sqrt{(2\pi)^n\det \mathbf \Sigma_k }}\sqrt{\frac{(2\pi)^n}{\det \mathbf{K}}}\right) \\
&= -\frac2\gamma \log \left(\frac{\exp\left(\frac{1}{2}\mathbf{y}_k^\top \mathbf{K}\mathbf{y}_k\right)}{\sqrt{\det (\mathbf \Sigma_k \mathbf{K})}}\right)\\
&= \frac1\gamma \log\det (\mathbf \Sigma_k \mathbf{K}) - \frac1\gamma \mathbf{y}_k^\top \mathbf{K}\mathbf{y}_k \\
&=\frac{1}{\gamma} \log \operatorname{det}\left(\mathbf{I}+\gamma Q \mathbf{\Sigma}_{k}\right) - \frac1\gamma \mathbf{y}_k^\top \mathbf{K}\mathbf{y}_k.
\end{aligned}
\end{equation}
By substituting $\mathbf{y}_k = -\gamma \mathbf{K}^{-1}Q\overline{\mathbf{X}}_k$, and $\mathbf{K} = \gamma Q + \mathbf\Sigma_k^{-1}$ into \eqref{eq:step6'}, we can simplify the second term  $-\frac1\gamma \mathbf{y}_k^\top \mathbf{K}\mathbf{y}_k$ as follows:
\begin{equation}\label{eq:step7}
-\frac1\gamma \mathbf{y}_k^\top \mathbf{K}\mathbf{y}_k = 
- \gamma \overline{\mathbf{X}}_k^\top Q \left(\gamma Q + \mathbf \Sigma^{-1}_k\right)^{-1} Q  \overline{\mathbf{X}}_k,
\end{equation}
assuming that $Q$ is a diagonal matrix and $\mathbf \Sigma_k$ is a symmetric matrix. 
Afterward, we can utilize the Woodbury matrix identity to simplify the expression $\left(\gamma Q + \mathbf{\Sigma}_k^{-1}\right)^{-1}$, where the Woodbury matrix identity is defined as \cite{woodbury1950inverting}:
\begin{equation}\label{eq:step8}
(A + UCV)^{-1} = A^{-1} - A^{-1}U(C^{-1} + VA^{-1}U)^{-1}VA^{-1}.
\end{equation}
By applying the Woodbury identity expressed in \eqref{eq:step8} to the given expression with $A = \gamma Q$, $C = \mathbf{\Sigma}_k^{-1}$, and $U = V = \mathbf{I}$ (where $\mathbf{I}$ denotes the identity matrix), we can rewrite $\left(\gamma Q + \mathbf{\Sigma}_k^{-1}\right)^{-1}$ as follows:
\begin{equation}\label{eq:step9}
\left(\gamma Q + \mathbf \Sigma_k^{-1}\right)^{-1} = 
\frac{1}{\gamma} Q^{-1} \Bigr[\mathbf{I} - Q_{\mathrm{rs}}Q^{-1}\Bigl], 
\end{equation}
where $Q_{\mathrm{rs}} = {\left(Q^{-1}+\gamma \mathbf{\Sigma}_{k}\right)^{-1}}$.
By substituting \eqref{eq:step9} back into \eqref{eq:step7}, we obtain:
\begin{equation}\label{eq:step10}
-\frac1\gamma \mathbf{y}_k^\top \mathbf{K}\mathbf{y}_k = 
-\left\|\overline{\mathbf{X}}_k\right\|_{Q}^{2}+ \left\|\overline{\mathbf{X}}_k\right\|_{Q_{\mathrm{rs}}}^{2}.
\end{equation}
Subsequently, by substituting \eqref{eq:step10} into \eqref{eq:step6'} and further substituting the obtained results into \eqref{eq:step2}, while replacing $\overline{\mathbf{X}}_k$ with $\Bar{\mathbf{x}}_k - \mathbf{x}_f$, the desired result can be obtained as follows:
\begin{equation}\label{eq:Q_rs_final}
q_{\text{rs}}(\mathbf{x}_k) = \frac{1}{\gamma} \log \operatorname{det}\left(\mathbf{I}+\gamma Q \mathbf{\Sigma}_{k}\right)
+\left\|\mathbf{\Bar{\mathbf{x}}}_{k}-\mathbf{x}_{f}\right\|_{Q_{\mathrm{rs}}}^2.
\end{equation}
\ihab{Remarkably, as \(\gamma\) approaches zero, the term \(\frac{1}{\gamma} \log \operatorname{det}(\mathbf{I}+\gamma Q \mathbf{\Sigma}_{k})\) approximates \(\operatorname{Tr}(Q \mathbf{\Sigma}_{k})\), avoiding the singularity that occurs if \(\gamma = 0\). Therefore, as $\gamma \to 0$, \( q_{\text{rs}}(\mathbf{x}_k) \) simplifies to:
\[
q_{\text{rs}}(\mathbf{x}_k) = \operatorname{Tr}(Q \mathbf{\Sigma}_{k})+\left\|\mathbf{\Bar{\mathbf{x}}}_{k}-\mathbf{x}_{f}\right\|_{Q_{\mathrm{rs}}}^2, 
\] where $Q_{\mathrm{rs}} = Q$ and \(\operatorname{Tr}(\cdot)\) denotes the trace of a matrix.}
It is worth emphasizing that a similar form of \eqref{eq:Q_rs_final}, along with its derivation, can be found in \cite{hyeon2020fast}. However, the presented derivation process in \cite{hyeon2020fast} is brief and lacks detailed explanations and clarifications.

\end{document}